\pdfoutput=1

\documentclass[11pt]{article}

\usepackage{acl}

\usepackage{times}
\usepackage{latexsym}

\usepackage[T1]{fontenc}

\usepackage[utf8]{inputenc}

\usepackage{microtype}

\usepackage{tabularx}
\usepackage{amsfonts}
\usepackage{amsmath}
\usepackage{booktabs}
\usepackage{xspace}
\usepackage{multirow}
\usepackage{algorithm}
\usepackage{algpseudocode}
\usepackage{graphicx}
\usepackage{CJKutf8}
\usepackage{mathrsfs}
\usepackage{bbm}
\usepackage{enumitem}
\usepackage{color}
\usepackage{arydshln}
\usepackage{subcaption}
\usepackage[switch]{lineno}

\newcommand{\nop}[1]{}
\newcommand{\add}[1]{\textcolor{red}{#1}}

\DeclareMathOperator*{\argmax}{arg\,max}
\DeclareMathOperator*{\argmin}{arg\,min}

\newcommand{\func}{semantic perturbation functions\xspace}
\newcommand{\attack}{\texttt{SemAttack}\xspace}

\newcommand{\m}[1]{{\textcolor{black}{{#1}}}}

\title{\attack: Natural Textual Attacks via Different Semantic Spaces}

\author{%
  \thanks{$\;$ Equal Contribution}$\,$ Boxin Wang$^1$, $^*$Chejian Xu$^1$, Xiangyu Liu$^2$, Yu Cheng$^3$, Bo Li$^1$ \\
  $^1$University of Illinois at Urbana-Champaign \\
  $^2$Alibaba Group,
  $^3$Microsoft Research \\
  \texttt{\small\{boxinw2,chejian2,lbo\}@illinois.edu}  \\
  \texttt{\small eason.lxy@alibaba-inc.com, yu.cheng@microsoft.com} \\ 
}

\begin{document}

\maketitle

\begin{abstract}

Recent studies show that pre-trained language models (LMs) are vulnerable to textual adversarial attacks.
However, existing attack methods either suffer from low attack success rates or fail to search efficiently in the exponentially large perturbation space. 
We propose an efficient and effective  framework \attack to generate natural adversarial text by constructing different \func.
In particular, \attack  optimizes the generated perturbations constrained on generic semantic spaces, including typo space, knowledge space (\emph{e.g.}, WordNet), contextualized semantic space (\emph{e.g.}, the embedding space of BERT clusterings), or the combination of these spaces.
Thus, the generated adversarial texts are more semantically close to the original inputs.
Extensive experiments reveal that state-of-the-art (SOTA) large-scale LMs (\emph{e.g.,} DeBERTa-v2)  and defense strategies (\emph{e.g.,} FreeLB) are still vulnerable to \attack.
We further demonstrate that \attack is general and able to generate natural adversarial texts for different languages (\emph{e.g.,} English and Chinese) with high attack success rates. 
Human evaluations also confirm that our generated adversarial texts are natural and barely affect human performance. Our code is publicly available at \url{https://github.com/AI-secure/SemAttack}.

\end{abstract}

\section{Introduction}

\nop{
\nop{Pretrained language representation models, including BERT \citep{Devlin2019BERTPO} and XLNet \citep{Yang2019XLNetGA}, have obtained state-of-the-art results over many downstream NLP tasks.}
\add{Recently, the impressive performance of BERT~\citep{Devlin2019BERTPO} has inspired many pre-trained large-scale language models \citep{Yang2019XLNetGA}~\pan{add more citations}, which have obtained state-of-the-art results over many downstream NLP tasks.}
\citet{Tenney2019BERTRT} points out BERT-based models can disambiguate information from high-level representation, which makes them effective in ambiguous languages 
\sout{ (such as Chinese). In Chinese, there are no explicit word delimiters, and the granularity of words is less well defined than other languages (such as English) \citep{ding-etal-2019-neural}}.
Moreover, \citet{li-etal-2019-word-segmentation} find that character-based models (\textit{e.g.} BERT) outperform word-based models \sout{that} \add{in the Chinese environment since the latter} are vulnerable to data sparsity and out-of-vocabulary \add{situations}\nop{in the Chinese environment}. \add{In Chinese, there are no explicit word delimiters, and the granularity of words is less well defined than other languages (such as English) \citep{ding-etal-2019-neural}.}}

Deep neural networks have achieved remarkable success in many machine learning tasks. Particularly, BERT~\citep{Devlin2019BERTPO} has inspired a suite of large-scale pre-trained language models \citep{Yang2019XLNetGA, zhang-etal-2019-ernie,Lan2019ALBERTAL}, which achieved new SOTA for many NLP tasks. In addition to BERT's dominant performance on English datasets, \citet{Tenney2019BERTRT} points out that BERT is similarly effective on other languages such as Chinese, whose granularity of words is more complex, given the model's ability to disambiguate information from high-level representations \citep{ding-etal-2019-neural}.

\begin{table}[t]\small \setlength{\tabcolsep}{7pt}
\centering
\begin{CJK*}{UTF8}{gbsn}
\resizebox{0.48\textwidth}{!}{
\begin{tabular}{p{7.8cm}}
\toprule 
\textbf{Original Input:} They need to hire \textcolor{blue}{experienced} sales rep who are mature enough to handle questions and sales.  \\
\textbf{Adversarial Input: } They need to hire \textcolor{red}{skilled} sales rep who are mature enough to handle questions and sales. \\ \hdashline[1pt/2pt]
\textbf{Sentiment Prediction: }  \textcolor{blue}{Most Negative} $\rightarrow$  \textcolor{red}{Most Positive} \\
\midrule
\textbf{Original Input: }  {\scriptsize 拿 \textcolor{blue}{什}么 能 吸 引 你 ： 我 们 的 海 外 学 子 ？} \\
(\textbf{Translation:} 
 \textcolor{blue}{What} can attract you: our overseas students? ) \\
\textbf{Adversarial Input: } {\scriptsize 拿 \textcolor{red}{甚}么 能 吸 引 你 ： 我 们 的 海 外 学 子 ？}
 \\
(\textbf{Translation: }\textcolor{red}{What} can attract you: our overseas students?)  \\  \hdashline[1pt/2pt]
\textbf{Topic Prediction: } \textcolor{blue}{Education News} $\rightarrow$ \textcolor{red}{Entertainment News} \\
\bottomrule
\end{tabular}
}
\end{CJK*}
\caption{{\small Adversarial texts generated against  English and Chinese BERT classifiers by \attack on Yelp and THUCTC datasets. Replacing \textcolor{blue}{a word/character} with \textcolor{red}{an adversarial one}  misleads the 
\textcolor{blue}{correct prediction} to a \textcolor{red}{wrong class} without fooling human.}}
\label{tab:example}
\end{table}

\begin{figure*}
    \centering
    \includegraphics[width=\linewidth]{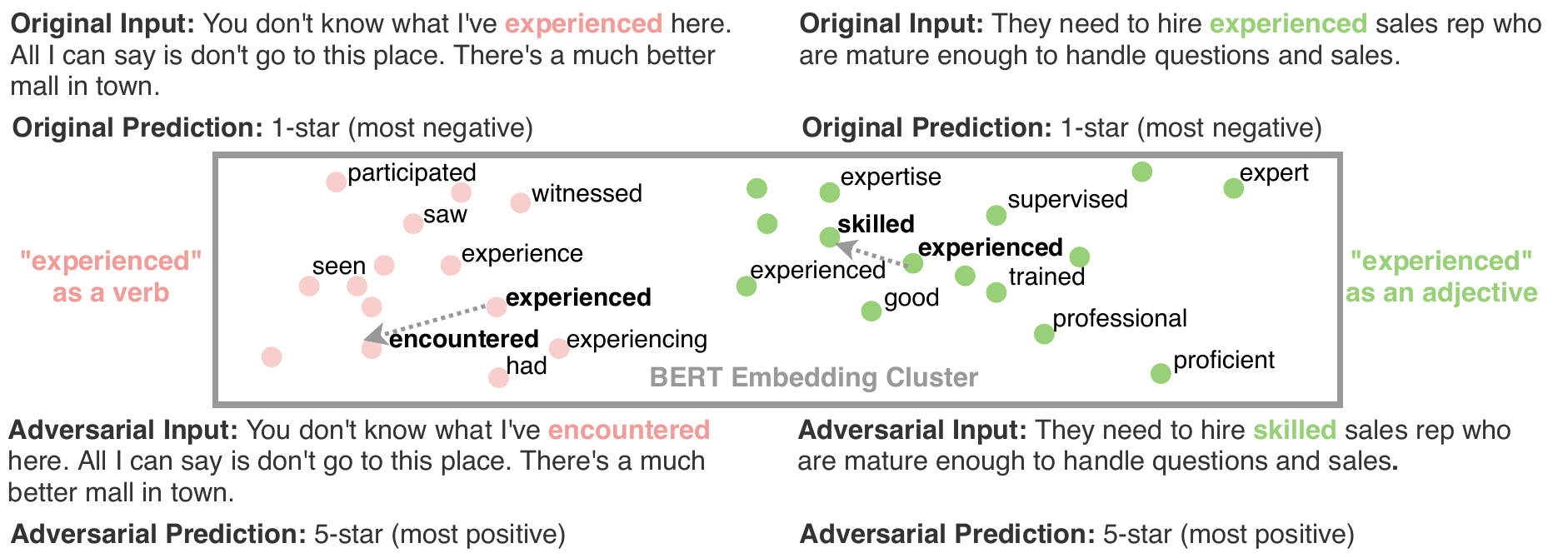}
    \caption{\small Adversarial texts against BERT sentiment classifier generated by \attack that formulates two different contextualized semantic perturbation spaces based on BERT embedding clusters (the embedding space is projected by PCA onto 2D space). The word ``experienced'' reveals different meanings (past tense of the verb ``experience'' or adjective form) in different contexts (clusters). Our contextualized semantic perturbation chooses ``saw'' or ``encountered'' as the perturbation for verb ``experienced'', while ``skilled'' or ``trained'' for the adjective form.}
    \label{fig:example}
\end{figure*}

Although effective for many NLP tasks, the robustness of these neural models is often challenged by carefully crafted adversarial examples. Specifically, attackers can add subtle human-imperceptible perturbation to the original input and induce dramatic changes in model output. 
Current adversarial text generation \citep{jia-liang-2017-adversarial, Li2018TextBuggerGA,alzantot-etal-2018-generating} is mainly heuristic and only achieves low attack success rates for BERT-based models. Other work \citep{DBLP:conf/aaai/ChengYCZH20, DBLP:conf/acl/EbrahimiRLD18} allows an input word to be substituted by any other word in the vocabulary, which fails to consider the semantic perturbation constraints and is prone to invalid adversarial examples. Recent work \citep{Jin2019IsBR, DBLP:conf/acl/ZangQYLZLS20} relies on external knowledge to constrain the perturbation yet poorly handles large search space that grows exponentially with the input length, as it requires hundreds of queries to generate one adversarial example in practice.

Furthermore, most existing textual adversarial attacks are not generalizable to other languages, due to unique language-dependent characteristics and the lack of universal linguistic resources. 
Moreover, character-level adversarial attacks designed in English context \citep{ebrahimi2017hotflip} are often ineffective for Chinese-character-level attacks, as the size of candidate characters increases by two orders of magnitude, resulting in surging computational costs especially for BERT-based models. 

We tackle these limitations in textual adversarial attacks by proposing an effective and efficient framework \attack, which can be used to further evaluate the robustness of NLP models. We generalize existing word-level attacks and propose generic semantic perturbation functions, which optimize and constrain the perturbations within different semantic spaces, so that the generated adversarial texts retain their semantic meaning. We mainly consider three types of
semantic spaces: ($1$) \emph{Typo Space}, using typo words or characters that can fool the models but not human judges;  ($2$) \emph{Knowledge Space}, utilizing external linguistic knowledge base (\emph{e.g.}, WordNet \citep{wordnet}) as valid perturbation candidates; and ($3$) \emph{Contextualized Semantic Space}, exploiting the embedding space of BERT to generate a contextualized perturbation set semantically close to the original word (Figure \ref{fig:example}).
The contextualized semantic space does not require additional knowledge, and therefore can scale to other languages, especially low-resource languages where a large knowledge base is unavailable.

After the candidate semantic space is determined, \attack searches for the optimal perturbation combination. Instead of requiring thousands of queries to generate one adversarial example, optimal perturbations can be efficiently found in the embedding space by solving an optimization problem. We also control the magnitude of perturbation to be small as shown in Table \ref{tab:example}. 
Extensive experiments on four datasets demonstrate that SOTA LMs and defense methods are still vulnerable to our adversarial attack, which are natural and barely affects human judgment. For example, the accuracy of BERT sentiment classifier drops from $70.6\%$ to $2.4\%$ by simply replacing fewer than $5\%$ words with our method. Although these adversarial examples are generated in the whitebox setting, they can effectively transfer to \m{two different} blackbox attack \m{settings} while retaining higher than $90\%$ attack success rate for BERT \m{and other large-scale LMs such as DeBERTa-XXLarge}. 

\underline{Our contributions} are summarized as follows: $1$) We propose a unified and effective adversarial attack framework \attack by constructing \func, which constraint perturbations within different semantic spaces and their combinations.
$2$) \attack generates contextualized perturbations that require no external knowledge and thus can easily adapt to different languages.
$3$) We conducted extensive experiments on different datasets and languages to show that adversarial texts generated by \attack are more semantically close to the benign inputs, and achieve much higher attack success rates than existing attack algorithms in different settings. 
$4$) Comprehensive studies demonstrate that SOTA LMs and defenses are still vulnerable to \attack, and human evaluation verifies the naturalness and validity of our adversarial examples.

\setlength{\belowdisplayskip}{1pt} \setlength{\belowdisplayshortskip}{2pt}
\setlength{\abovedisplayskip}{1pt} \setlength{\abovedisplayshortskip}{2pt}
\section{\attack}

\subsection{Problem Formulation}

Given an input $\boldsymbol{x}= [{x_0, x_1, ..., x_n}]$,
where 
$x_i$ is the $i$-th input token, 
the classifier $f$ maps the input to final logits $\boldsymbol{z} = f(\boldsymbol{x}) \in \mathbb{R}^{C}$, where $C$ is the number of classes, and outputs a label $y = \argmax f(\boldsymbol{x})$. 

During attack, we evaluate the effectiveness of attack algorithms by calculating the targeted attack success rate (TSR):
\begin{equation}
\footnotesize
    \text{TSR} = \frac{1}{|D_{\text{adv}}|} \sum_{\boldsymbol{x'} \in D_{\text{adv}}}{\mathbbm{1}[\argmax f(\boldsymbol{x'}) \equiv {y^*}]}
    \label{eq:tsr}
\end{equation}
and untargeted attack success rate (USR):
\begin{equation}
\footnotesize
    \text{USR} = \frac{1}{|D_{\text{adv}}|} \sum_{ \boldsymbol{x'} \in D_{\text{adv}}}{\mathbbm{1}[\argmax f(\boldsymbol{x'}) \neq {y}]}
    \label{eq:usr}
\end{equation}
where the attack algorithm generates one adversarial sentence for each sample to form an adversarial dataset $D_{\text{adv}}$ 
, $y^*$ is the targeted false class, $y$ is the ground truth label, and $\mathbbm{1}(\cdot)$ is the indicator function.

\subsection{Semantic Perturbation Functions}
To control adversarial examples to be semantically close to the original input, we design a general form of semantic perturbation function $\mathcal{F}$, which takes one token $x$ as input, and returns its candidate perturbation space $\mathcal{S}=\{x^*_0, x^*_1, ..., x^*_n\}$. We next discuss the types of perturbation function $\mathcal{F}$.

\begin{CJK*}{UTF8}{gbsn}
\noindent \textbf{Typo-based Perturbation Function $\mathcal{F}_T$}
constrain the search space $\mathcal{S}$ in the \emph{typo space}, which uses typo words or characters to replace original tokens so that human can still understand the original meaning while models are fooled.
In English, we follow the generation process introduced in TextBugger \cite{Li2018TextBuggerGA} to generate typos. 

In order to illustrate how our proposed method can be easily adapted to multilingual settings, we also generate typo-based semantic space for Chinese. Specifically, for each Chinese token $x$, we prepare a set of common Chinese characters $\mathcal{S}$ that look similar ({\scriptsize ``形近字''}) or have the same pronunciation ({\scriptsize ``音近字''}) as the original token $x$. We use the open-source similar Chinese character list that contains more than 9,000 common Chinese characters. To search for the Chinese characters with the same pronunciation (\textit{i.e.,} pinyin), we first query the pronunciation of input $x$ and then choose the characters returned based on the same pronunciation. If $x$ is a heteronym that has multiple pronunciations, we only use one pronunciation to do the query. We also limit the size of Chinese characters of the same pronunciation to be less than 6 so that the search space is not too large. For the Chinese example shown in Table \ref{tab:example}, we use {\scriptsize ``甚''} to replace {\scriptsize{``什''}} as they share the same pronunciation and are a common typo that will not affect human understanding.  
\end{CJK*}

\noindent \textbf{Knowledge-based Perturbation Function $\mathcal{F}_K$} considers the \emph{knowledge space} to constrain the perturbation search space $\mathcal{S}$. Specifically, $\mathcal{F}_K$ utilizes existing knowledge base to build a candidate perturbation set. 
In our work, we use WordNet as an example to illustrate how our framework can integrate rule-based knowledge to enhance the quality of adversarial examples. WordNet is a large lexical dataset of more than 200 languages that groups words into sets of cognitive synonyms. With the manually labeled semantic relations among words, synonyms queried from WordNet (\textit{i.e.}, synsets) share the same semantic meaning as the query word $x$. Therefore we choose these synonyms returned from WordNet to be the search space $\mathcal{S}$. 
We note that WordNet also contains hypernyms and hyponyms information, but including them into the search space may incur some unnatural replacement (\textit{e.g.}, replacing ``fifth'' with ``rank''). Therefore, we only consider synsets as the candidate search space $\mathcal{S}$. In addition, even for the same token (\textit{e.g}., ``use'') in WordNet, it may have different part-of-speech (POS) tags (\emph{e.g.}, ``use'' as verb or as noun), and thus has different synonyms (\emph{e.g.,} ``exploitation'' for noun ``use'' and ``practice'' as verb ``use''), which may result in nonsensical replacement. In order not to include synonyms that have unusual part of speech, we counted the frequency of POS in the synset and only selected the words with the most frequent POS. Using the synonym set $\mathcal{S}$ after filtering, we are able to generate adversarial input texts that mislead models' prediction while barely affect on human understanding.

\noindent \textbf{Contextualized Semantic Perturbation Function $\mathcal{F}_C$} is a novel perturbation function that explores the BERT embedding space and searches for contextualized perturbation to tackle the issue of most language tokens being polysemous. Previous work \citep{Li2018TextBuggerGA,Jin2019IsBR} takes it as a standard practice to use the proximity in embedding space to query the semantic similarity. However, their embedding space is built on a non-contextualized word embedding from GLoVE \citep{DBLP:conf/emnlp/PenningtonSM14} or Word2Vec \citep{DBLP:conf/nips/MikolovSCCD13}, thus failing to consider the polysemy when generating the perturbation. We propose to explore the BERT embedding space, which is verified by \citep{hewitt-manning-2019-structural, Coenen2019VisualizingAM, papadimitriou2021deep} that BERT embeddings can preserve  syntactic and semantic information for word sense disambiguation better than GLoVE or Word2Vec. So the contextualized space from $\mathcal{F}_C$ is valid semantic perturbations. Similar to our parallel work \citep{li2020bert} of using BERT to generate adversarial perturbations, $\mathcal{F}_C$ also does not require external linguistic resources such as POS checker. Thus $\mathcal{F}_C$ can be adapted to other languages, as long as pre-trained BERT of such language models is available.

Specifically, we first choose a set of commonly used tokens $\mathcal{X}$. For each word $x \in \mathcal{X}$, we select at most 100 example sentences from Wikipedia that contain the word $x$ so that these sentences represent different meanings of $x$ in different contexts. We then feed these sentences into a pretrained BERT model to obtain the contextualized embeddings for each word $x$. Finally, the contextualized embeddings for all words in $\mathcal{X}$ formulate a large BERT embedding space.  Figure \ref{fig:example} visualizes a BERT embedding space projected into 2D space by PCA.

To query the search space $\mathcal{S}$ for token $x$, we first calculate the BERT embedding of token $x$ given its context sentence. Even for the same token, given different contexts and meanings, BERT will generate distinct representations in the high dimensional embedding space. For the example in Figure 1, the token ``experienced'' given different contexts have different latent representations and neighbors. Then we use $k$ nearest neighbors (KNN) algorithm to choose the neighbors of the contextualized embedding of $x$ as its perturbation search space $\mathcal{S}$. To ensure high quality of search space $\mathcal{S}$, we further filter $\mathcal{S}$ and only return the words that appear more frequently than a threshold $\epsilon$ among $k$ nearest neighbors. In this way, we remove the noisy tokens that are rarely used and  retain the high-quality neighbor tokens whose contextualized semantics are mostly close to the original token $x$.

\noindent \textbf{Discussion.} The final search space $\mathcal{S}$ can be the union of the search spaces mentioned above. This makes existing defense algorithm~\citep{DBLP:conf/acl/JonesJRL20}  difficult to apply, as they can only defend against typo-based perturbation but fails to detect other types of perturbation.

$\mathcal{F}$ is a generalization of most existing word-level textual adversarial attacks. 
Though $\mathcal{F}_T$ and $\mathcal{F}_K$ have been discussed in the previous literature (see \S Related Work),
we note that the goal of our paper is not to improve or propose better typo or knowledge perturbation, but to consider multiple semantic spaces at the same time  
to help generate natural high-quality adversarial examples. 

\subsection{Attack Algorithm \attack}

The full pipeline is shown in \m{Appendix} Algorithm \ref{algo}. Essentially, \attack searches for the optimal perturbations from different semantic spaces determined by semantic perturbation functions, which is efficiently solved as an optimization problem so that we only perturb as few tokens as possible while achieving the targeted attack. 

Unlike generating adversarial examples in the continuous data domain, it is difficult to directly utilize the gradient to guide token substitution due to the discrete nature of text. Thus, we search perturbation in the embedding space and map the perturbed embedding back to tokens. Specifically, the one-hot representation of each discrete token $\boldsymbol{x_i} \in \mathbb{R}^{|V|}$ ($V$ is the vocabulary set) is mapped into an embedding space of dimension $d_c$ via the embedding matrix $\boldsymbol{M_e} \in \mathbb{R}^{d_c\times |V|}$
\begin{equation} \small
    [\boldsymbol{e_1; e_2; ...; e_n}] = \boldsymbol{M_e}\Big[\boldsymbol{x_0; x_1; ...; x_n}\Big].
\end{equation}
We optimize perturbation $\boldsymbol{e^*}$ added to the original embedding $\boldsymbol{e}$ for $m$ iterations. In each iteration, we freeze all the parameters of the classifier $f$ and optimize variable $\boldsymbol{e^*}$ only. Following \citet{Carlini2016TowardsET}, we minimize the loss function as:
\begin{equation}\small
    \mathcal{L}(\boldsymbol{e^*})  =  ||\boldsymbol{e^*}||_p + c \cdot g(\boldsymbol{x'}),
    \label{cw}
\end{equation}
where the first term 
controls the magnitude of perturbation, while $g(\cdot)$ is the attack objective function depending on the attack scenario. $c$ weighs the attack goal against attack cost. 

In \emph{targeted attack} scenarios, we define $g(\cdot)$ as:
\begin{equation*}\small
    g(\boldsymbol{x'}) = \text{max}[\text{max}\{f(\boldsymbol{x'})_i:i \neq t\} - f(\boldsymbol{x'})_t, -\kappa],
\end{equation*}
where $t$ is the targeted false class and $f(\boldsymbol{x'})_i$ is the $i$-th class logit. A larger $\kappa$ encourages the classifier to output targeted false class with higher confidence. 

In \emph{untargeted attack} scenarios, $g(\cdot)$ becomes
\begin{equation*}\small
    g(\boldsymbol{x'}) = \text{max}[f(\boldsymbol{x'})_t - \text{max}\{f(\boldsymbol{x'})_i:i \neq t\} , -\kappa],
\end{equation*}
where $t$ is the ground truth class. 

After each iteration of gradient descent, we have an optimized perturbation $\boldsymbol{e^*}$ in the embedding space that tends to fool the classifier $f$ with small perturbations. We choose the perturbed token $\boldsymbol{x'_i} \in \mathcal{S}=\mathcal{F}(x_i)$ that is from the semantic search space $\mathcal{S}$ returned by  $\mathcal{F}(x_i)$ and semantically closest to the perturbed embedding $\boldsymbol{e'_i}$.
\begin{equation}\small
\begin{aligned}
    & \boldsymbol{e'_i} = \boldsymbol{e_i} + \boldsymbol{e^*_i}, \\
    & \boldsymbol{x'_i} = \argmin_{\boldsymbol{x'_i} \in \mathcal{S}} (  || \boldsymbol{e’_i} - \boldsymbol{M_e x’_i} ||_p).
\end{aligned}
\end{equation}

Finally, we obtain an optimal perturbation $\boldsymbol{e^*}$ after repeating the optimization step and token substitution step for $m$ iterations. Under such settings and constraints, most tokens remain the same and very few are perturbed to their semantically close neighbors. Thus, the adversarial examples still look valid to humans but can fool the models.

\section{Experimental Results}
\label{sec:exp}

In this section,
\m{we conduct comprehensive experiments to evaluate our attack method in various settings. We \textbf{first}} apply our attack method to two standard NLP models, BERT and SOTA Self-Attention LSTM. We evaluate on \textit{two different types of NLP tasks}, sentiment analysis and natural language inference (NLI). 
\m{\textbf{Secondly}, we investigate the effectiveness of \attack against SOTA large-scale language models and defense methods. \textbf{Thirdly}, we take Chinese as an example to measure \attack's generalization ability across different languages.} We evaluate BERT models finetuned on two Chinese datasets. 
\m{\textbf{Finally}, we conduct extensive human evaluations on both English and Chinese datasets.}

\m{We find that: 
$1$) \attack can achieve better attack success rates than existing textual adversarial attack methods with better language quality and adversarial transferability.
$2$) SOTA LMs and defense methods are still vulnerable to  our \attack. 
$3$) \attack is a general textual adversarial attack framework and can be easily adapted to other languages in addition to English with high attack success rates.
$4$) Adversarial examples generated by \attack are natural and barely affect human performance.
}

\begin{table*}[t]\small \setlength{\tabcolsep}{7pt}
\centering
\begin{subtable}[t]{.48\textwidth}
\centering
\resizebox{1.0\linewidth}{!}{
\begin{tabular}{ccccc}
\toprule
Model & Attack Method & \% USR/TSR  & \% Perturbation\\
\midrule
\multirow{5}{*}{\shortstack{BERT\\(Acc: \\0.706)}}
& HotFlip   & 71.5/24.0 & 14.9/44.9\\
& \attack (+$\mathcal{F}_T$)   & 42.4/9.3 & 4.7/9.1\\
& \attack (+$\mathcal{F}_K$)   & 84.6/69.3 & 6.7/13.9\\
& \attack (+$\mathcal{F}_C$)  & 91.3/79.7 & 4.7/11.1\\
& \attack (+all)  & \textbf{97.6}/\textbf{93.8} & {4.3/10.2}\\
\midrule
\multirow{5}{*}{\shortstack{Self-Attention \\LSTM\\(Acc:\\ 0.705)}}
& HotFlip   & 16.3/3.2 & 2.5/17.4\\
& \attack (+$\mathcal{F}_T$)  & 67.2/49.4 & 14.7/21.1\\
& \attack (+$\mathcal{F}_K$)  & 47.9/43.6 & 10.4/18.3\\
& \attack (+$\mathcal{F}_C$)  & 67.3/56.5 & 15.1/23.2\\
& \attack (+all)  & \textbf{88.1}/\textbf{84.0} & 19.2/29.2\\
\bottomrule
\end{tabular}
}
\caption{Yelp Dataset}
\label{tab:english_yelp}
\end{subtable}
\quad
\begin{subtable}[t]{.48\textwidth}
\centering
\resizebox{1.0\linewidth}{!}{
\begin{tabular}{ccccc}
\toprule
Model & Attack Method & \% USR/TSR & \% Perturbation\\
\midrule
\multirow{5}{*}{\shortstack{BERT\\(Acc: \\0.829)}}
& HotFlip   & 83.3/44.9 & 27.0/30.3\\
& \attack (+$\mathcal{F}_T$)   & 21.2/10.2 & 13.1/16.5\\
& \attack (+$\mathcal{F}_K$)   & 53.8/23.2 & 14.8/22.3\\
& \attack (+$\mathcal{F}_C$)  & 90.2/69.7 & 15.3/26.9\\
& \attack (+all)  & \textbf{92.6}/\textbf{72.6} & 15.6/20.0\\
\midrule
\multirow{5}{*}{\shortstack{Self-Attention \\LSTM\\(Acc:\\ 0.705)}} 
& HotFlip   & 32.3/17.8 & 11.6/13.4\\
& \attack (+$\mathcal{F}_T$)   & 53.8/33.4 & 23.9/29.1\\
& \attack (+$\mathcal{F}_K$)   & 40.7/23.2 & 21.4/22.2\\
& \attack (+$\mathcal{F}_C$)  & 76.5/63.8 & 30.9/36.3\\
& \attack (+all)  & \textbf{86.2}/\textbf{68.5} & 39.0/36.9\\
\bottomrule
\end{tabular}
}
\caption{SNLI Dataset}
\label{tab:english_snli}
\end{subtable}
\caption{
\small Whitebox attack success rate for different attacks under targeted/untargeted attacks (TSR/USR) and corresponding word perturbation percentage against self-attention LSTM and BERT on Yelp and SNLI datasets.} 
\label{tab:english}
\end{table*}

\begin{table}[t]\small \setlength{\tabcolsep}{7pt}
\centering
\resizebox{1.0\linewidth}{!}{
\begin{tabular}{cccc}
\toprule
Model & Attack Method & \% USR/TSR & \% Perturbation\\
\midrule
\multirow{6}{*}{\shortstack{DeBERTa\\(Large,\\Acc: 0.928)}}& TextFooler & 83.2/57.1 & 22.5/21.3 \\
& BERT-ATTACK & 84.4/36.6 & 19.4/17.9 \\
& \attack(+$\mathcal{F}_T$) & \textbf{88.1}/\textbf{58.3} & 17.8/16.0 \\
& \attack(+$\mathcal{F}_K$) & 82.1/53.7 & 22.1/20.9 \\
& \attack(+$\mathcal{F}_C$) & 80.3/33.6 & 27.6/27.7 \\
& \attack(+all) & 83.0/41.2 & 21.4/20.5 \\
\midrule
\multirow{6}{*}{\shortstack{DeBERTa\\(XXLarge-v2,\\Acc: 0.931)}}& TextFooler & 86.4/57.1 & 22.1/20.3 \\
& BERT-ATTACK & 83.4/37.2 & 19.2/17.8 \\
& \attack(+$\mathcal{F}_T$) & \textbf{90.5}/\textbf{65.5} & 17.6/16.2 \\
& \attack(+$\mathcal{F}_K$) & 86.8/58.4 & 22.3/21.7 \\
& \attack(+$\mathcal{F}_C$) & 80.6/38.7 & 27.6/27.9 \\
& \attack(+all) & 82.7/42.9 & 21.2/20.2 \\
\midrule
\multirow{6}{*}{\shortstack{FreeLB\\(Acc: 0.924)}} & TextFooler & 63.0/31.5 & 22.1/22.0 \\
& BERT-ATTACK & 65.6/31.1 & 19.1/18.6 \\
& \attack(+$\mathcal{F}_T$) & \textbf{71.4}/26.2 & 17.0/14.7 \\
& \attack(+$\mathcal{F}_K$) & 63.2/32.6 & 22.9/23.9 \\
& \attack(+$\mathcal{F}_C$) & 66.7/\textbf{32.7} & 27.8/28.0 \\
& \attack(+all) & 64.3/32.2 & 20.9/20.5 \\
\bottomrule
\end{tabular}
}
\caption{
\small \m{Zero-query blackbox attack success rate for different attacks under targeted/untargeted attacks (TSR/USR) and corresponding word perturbation percentage against large-scale LMs and defense methods on SNLI datasets.}}
\label{tab:large_models}
\end{table}

\subsection{
Whitebox and Blackbox Attack
}
\label{sec:main}
\paragraph{Datasets}
\m{For sentiment classification task, we choose the standard 5-class sentiment classification dataset, Yelp dataset. Note that unlike previous work \citep{li2020bert,Jin2019IsBR} that uses binary sentiment classification dataset, we focus on the standard 5-class Yelp dataset to further evaluate the \textbf{targeted attack capability} of \attack.} For NLI task, we choose SNLI dataset. The detailed dataset descriptions are in Appendix \m{\S\ref{appendix:dataset}}.

\paragraph{Models}
We evaluate the robustness of \textit{BERT} and \textit{Self-Attention LSTM} \citep{lin2017structured}. We present their test accuracy on the benign test sets in Table \ref{tab:english}.
More hyperparameter settings and training details are discussed in Appendix \m{\S\ref{appdix:model}}.

\paragraph{Attack Baselines}
We consider SOTA whitebox and blackbox attack baselines.

\noindent $\bullet \;$\textbf{HotFlip} \cite{ebrahimi2017hotflip} is a whitebox attack method for generating adversarial examples on both character-level and word-level. In terms of preserving semantic meaning, we only use word-level attacks in our experiments, which uses gradient-based optimization method to flip words.

\noindent $\bullet \;$\textbf{TextFooler} \cite{Jin2019IsBR} is a blackbox attack method for generating adversarial text, which uses similarities between pre-calculated word embeddings to find synonyms for each word. 

\noindent $\bullet \;$\textbf{BERT-Attack} \cite{li2020bert} is a strong blackbox attack method using pre-trained masked language models such as BERT to replace words in input sentences, where pre-trained masked language models provide candidate words that have high semantic similarity between original texts. 

These methods all perform untargeted attacks. We adapt them to both untargeted and targeted attack settings in our experiments.

\paragraph{Attack Goal} In the sentiment analysis task, we consider the \textbf{targeted attack}, and choose the most opposite sentiment class as the targeted class, so sentences with original label lower than $2$ (\textit{negative}) are attacked to class $4$ (\textit{most positive}), and others are attacked to class $0$ (\textit{most negative}). In the NLI task, \textit{Contradiction} and \textit{neutral} will be attacked to \textit{entailment} while \textit{entailment} will be attacked to \textit{contradiction}.

\paragraph{Adversarial Attack Evaluation}
We perform \attack on BERT and LSTM-based classifiers in both the whitebox and blackbox settings. 
The whitebox setting approximates the worst-case scenario, where attackers have the access to the model parameters and gradients; while the blackbox setting assumes that attackers can only access the model's output confidence. 

For the \textbf{whitebox attack} shown in Table \ref{tab:english}, \attack can outperform all the SOTA baselines and achieve the highest success rates in both untargeted and targeted settings for BERT and LSTM-based models with smaller or comparable perturbation rates. For example, untargeted \attack achieves $97.6\%$ attack success rate for BERT models by perturbing $4\%$ words on the Yelp dataset, when searching from the combination of the semantic spaces of $\mathcal{F}_T$, $\mathcal{F}_K$ and $\mathcal{F}_C$.

To adapt \attack to the blackbox attack setting, we distill the blackbox (teacher) model to train a whitebox (student) model, and \textit{transfer} the adversarial examples from the whitebox student model to attack the blackbox model. More details can be found in Appendix \S\ref{sec:exp_set}.

For the \textbf{blackbox attack} shown in Appendix Table \ref{tab:english_blackbox}, the transferability-based \attack achieves higher attack success rates than SOTA blackbox attacks for self-attention LSTM. 
We also observe that BERT-ATTACK achieves a higher attack success rate on BERT than \attack. We think it is mainly because that  BERT-ATTACK adopts an aggressively large candidate perturbation size (top-$k$=48), which may lead to large semantic changes (indicated by the worse human performance as shown in Table~\ref{tab:quality}). For instance, we observe that some words are even changed to their antonym in BERT-ATTACK. On the contrary, the average size of search spaces for \attack(+all) is only $11.87$, aiming to guarantee the naturalness and validity of the generated adversarial examples. We present more details of our semantic space in Appendix \S \ref{appendix:func}.

In addition, we observe that Self-Attention LSTM models are more robust than BERT in most settings. For example, we achieve the highest USR of $88.1\%$ in whitebox attack on the Yelp dataset, which is $9.5\%$ lower than BERT in the same setting. This suggests that self-attention mechanism can improve the robustness of vanilla WordLSTM by a large margin, as WordLSTM is known less robust than BERT \citep{Jin2019IsBR}. 

\subsection{
Attack SOTA LMs and Defense Methods
}
\m{In this section, we evaluate \attack and baseline attacks against various SOTA large-scale language models and defense methods.}

\paragraph{\m{Dataset and Attack Baselines}}
\m{Following \S \ref{sec:main}, we evaluate \attack on SNLI dataset. We choose the same blackbox attack methods, TextFooler and BERT-Attack, as our baselines.}

\paragraph{\m{Models}}
\m{We consider the following models and defense methods following the Adversarial GLUE Benchmark \cite{wang2021adversarial}. The selected large-scale models and defense methods not only represent SOTA performance on NLU tasks, but also achieve the highest robustness in the leaderboard.}

\noindent $\bullet \;$ \m{\textbf{DeBERTa \citep{he2020deberta}} improves BERT-based models by introducing disentangled attention mechanism and enhanced mask decoder, which is one of the best models in the GLUE leaderboard \cite{wang2018glue}. In our experiment, we use DeBERTa (Large) and DeBERTa (XXLarge-v2).}

\noindent $\bullet \;$ \m{\textbf{FreeLB \citep{zhu2019freelb}} is an adversarial training algorithm that defends adversarial attacks by adding  perturbations to word embeddings and minimizing the corresponding adversarial loss.}

\paragraph{\m{Attack Goal}}
\m{To demonstrate the model robustness in an approximately real-world scenario, we consider a \textbf{zero-query setting}, a more rigorous and common scenario that assumes the target models are not accessible during the attack phase.
Since we can not access the target model, we perform a transferability-based backbox attack. Specifically, we attack the selected language models and defense methods using adversarial SNLI texts generated by \attack against BERT classifier in \S \ref{sec:main}. }

\paragraph{\m{Adversarial Attack Evaluation}}
\m{We finetune the above models on the SNLI dataset and attack them using adversarial texts generated against BERT. The results are shown in Table \ref{tab:large_models}.}

\m{For the \textbf{zero-query setting}, \attack always achieves the highest success rates. Specifically, among all the attack methods, \attack(+$\mathcal{F}_T$) always has the highest USR regardless of the model it is tested on. For example, on the largest model, DeBERTa (XXLarge-v2), we achieve $90.5\%$ USR, which is $7.1\%$ higher than BERT-ATTACK.}

\m{Furthermore, we find that \textit{increasing the number of model parameters and expanding the model architecture have little effect on defense against adversarial attacks}. DeBERTa (XXLarge-v2), for example, is substantially larger than DeBERTa (Large), yet the attack success rates are similar. In some cases DeBERTa (XXLarge-v2) is even less robust than DeBERTa (Large). We also observe that introducing some defense strategies slightly improves the model's robustness. When we use the defense strategy of FreeLB, we can see that the robustness  increases, but it is still not satisfactory to defend existing adversarial attacks.}

\begin{table}[t]\small \setlength{\tabcolsep}{7pt}
\centering
\resizebox{1.0\linewidth}{!}{
\begin{tabular}{ccccc}
\toprule
Dataset & Setting & Attack Method & \% USR/TSR  & \% Perturbation\\
\midrule
\multirow{10}{*}{\shortstack{THUNews\\(Acc: \\0.818)}}
& \multirow{5}{*}{\shortstack{White-\\box\\Attack}}& HotFlip   & 81.4/40.4  & 21.7/27.9\\
& & \attack(+$\mathcal{F}_T$)  & 96.6/81.7 & 20.1/34.7\\
& & \attack(+$\mathcal{F}_K$)  & 15.6/3.6 & 16.1/17.4\\
& & \attack(+$\mathcal{F}_C$)   & 95.0/78.3 & 17.4/29.4\\
& & \attack(+all)  & \textbf{99.0}/\textbf{92.1} & 15.1/26.3\\ 
\cmidrule{2-5}
& \multirow{5}{*}{\shortstack{Black-\\box\\Attack}} & HotFlip & 44.3/10.0  & 15.4/10.8\\
& & \attack(+$\mathcal{F}_T$)  & 52.3/34.0 & 19.7/35.3\\
& & \attack(+$\mathcal{F}_K$)  & 8.4/1.3 & 12.7/13.1\\
& & \attack(+$\mathcal{F}_C$)   & 55.9/37.0 & 17.6/28.6\\
& & \attack(+all)  & \textbf{58.6}/\textbf{48.2} & 16.4/25.8\\
\midrule
\midrule
\multirow{10}{*}{\shortstack{Wechat\\(Acc:\\ 0.891)}} & \multirow{5}{*}{\shortstack{White-\\box\\Attack}}
& HotFlip   & 95.2/0.0 & 11.4/-\\
& &\attack(+$\mathcal{F}_T$)   & 86.0/88.3 & 7.2/12.4\\
& &\attack(+$\mathcal{F}_K$)   & 32.8/24.5 & 5.2/7.6\\
& &\attack(+$\mathcal{F}_C$)  & 96.8/96.4 & 5.8/9.4\\
& &\attack(+all)  & \textbf{98.7}/\textbf{98.0} & 4.6/8.7\\
\cmidrule{2-5}
& \multirow{5}{*}{\shortstack{Black-\\box\\Attack}} & HotFlip  & 21.7/0.0 & 8.9/-\\
& &\attack(+$\mathcal{F}_T$)   & 49.4/35.8 & 7.3/17.4\\
& &\attack(+$\mathcal{F}_K$)   & 19.5/11.7 & 4.0/7.7\\
& &\attack(+$\mathcal{F}_C$)  & 51.8/\textbf{42.4} & 5.3/12.2\\
& &\attack(+all)  & \textbf{54.5}/36.7 & 4.0/11.7\\
\bottomrule
\end{tabular}
}
\caption{
\small Whitebox and blackbox attack success rate for different attacks under targeted/untargeted attacks (TSR/USR) and corresponding word perturbation percentage against Chinese BERT on THUNews and Wechat Finance datasets.} 
\label{tab:chinese}
\end{table}

\subsection{Adapt \attack to Chinese}

\paragraph{Datasets}
We evaluate our performance on the following two datasets in Chinese\m{: 14-category news classification dataset THUNews \citep{sun2016thuctc} and $11$-class Wechat Finance dataset. More details about these datasets are introduced in Appendix \ref{appendix:dataset}}.

\paragraph{Models}
We use BERT pre-trained on Chinese corpora and finetune on the two datasets separately. 
After finetuning, our BERT achieved $0.818$ accuracy on THUNews dataset and $0.891$ on Wechat Finance Dataset, as shown in Table \ref{tab:chinese}

\paragraph{Attack Baselines}
\m{Since both TextFooler and BERT-Attack adopt an aggressively large perturbation candidate space and thus require additional language resources (e.g., POS checker; stop words filtering) to ensure the proposed candidate words are valid, they cannot be adapted to Chinese due to the lack of corresponding language resources. Therefore}, we adapt \textbf{HotFlip} for Chinese classification task, since it does not rely on any other linguistic resources.  We also adapt it to transferability-based blackbox attack settings as well as the targeted attack setting for fair comparison.

\paragraph{Attack Goal}
In this paper, we choose the targeted attack class as ``technology news'' for THUNews dataset and ``Bank'' for Wechat dataset (when the ground truth label is the targeted class, we switch the target to another random class). This strategy achieves the highest targeted attack success rate as shown in Appendix \m{\ref{sec:strategy}}. 

\paragraph{Adversarial Attack Evaluation} 

In the \textbf{whitebox attack} scenario in Table \ref{tab:chinese}, \attack is able to make the model mistakenly classify nearly all sentences with only a small number of characters being manipulated in both targeted and untargeted settings. The untargeted attack achieves $99\%$ success rate by substituting merely two tokens on average on the THUNews dataset. On Wechat Finance dataset, it achieves $98.7\%$ attack success rate by perturbing $4.6\%$ tokens on average in the input sequences. In the targeted attack scenario, we always make BERT output as our expected false class on both datasets, resulting in a huge performance drop on BERT models. We achieve $92.1\%$ and $98.0\%$ on THUNews dataset and Wechat Finance dataset, respectively.

We also present the \textbf{blackbox attack} results in Table \ref{tab:chinese}. We can see that \attack(+all) achieves the highest success rates in most cases, which suggests that our semantic perturbation spaces have high adversarial transferability. Note that we do not present the targeted attack on Wechat Finance dataset for HotFlip since all attack attempts failed.

\paragraph{Ablation Studies}
We conduct a series of ablation studies such as exploration of BERT embedding space, attack strategies, $\ell_p$ norm selection for Eq.(\ref{cw}), hyper-parameter selection, and attack efficiency comparison, etc. in Appendix \ref{appdix:analysis}.

\begin{table}[t] \setlength{\tabcolsep}{7pt}
\begin{small}
\centering
\resizebox{\linewidth}{!}{
\begin{tabular}{cccccc}
\toprule
Dataset & Attack Method & $\%$ Perturbation & PPL & BertScore & Human Ratings \\
\midrule
\multirow{4}{*}{\shortstack{Yelp\\(English)}} & HotFlip  & 14.9 & 57.1 & 0.79 & 3.337 $\pm$ 1.650 \\
& TextFooler  & 13.5 & 43.7 & 0.78 & 3.361 $\pm$ 1.326 \\
& BERT-ATTACK  & \textbf{4.2} & \textbf{31.4} & \textbf{0.92} & 3.513 $\pm$ 1.280 \\
& \attack(+all) & 4.3 & 34.4 & 0.91 & \textbf{3.524 $\pm$ 1.584} \\
\midrule 
\multirow{2}{*}{\shortstack{THUNews\\(Chinese)}} & HotFlip  & 21.7 & 488.3 & 0.60 & 3.770 $\pm$ 1.061 \\
& \attack(+all)  & \textbf{15.1} & \textbf{317.4} & \textbf{0.76} & \textbf{3.846 $\pm$ 0.906} \\
\bottomrule
\end{tabular}
}
\caption{Language quality evaluation for the generated adversarial texts in both Chinese and English.}
\label{tab:quality}
\end{small}
\end{table}

\subsection{Adversarial Text Quality Evaluation}
To confirm that our generated adversarial texts are valid and natural to humans, we conduct both \textit{automatic evaluation} and \textit{human evaluation} on both English and Chinese NLP tasks, considering \underline{language quality} and \underline{utility preservation}. More evaluation details can be found  in Appendix \ref{appendix:human}.

\paragraph{Language Quality Evaluation} We sample $100$ original sentences from the test set for both Chinese and English such that all of them can be successfully attacked by \attack and our baselines. 
For automatic evaluation, we consider the average perturbation rate, perplexity (PPL) (based on GPT-2), and BertScore as metrics to indicate the language quality.
For human evaluation, we present every generated adversarial sentence to 5 human annotators, ask them to rate the language quality from 1 to 5, and calculate the average ratings. We present the evaluation results in Table \ref{tab:quality}.

We can see that \attack has the best human ratings across different baselines for both Chinese and English. In terms of automatic evaluation metrics, we observe that \attack is quite close to the SOTA BERT-ATTACK. We think the reason why \attack is slightly weaker than BERT-ATTACK in terms of PPL and BertScore is that \attack also considers typos and knowledge-based perturbations. Such perturbations usually look good to humans, but may greatly impact the scores calculated by pretrained language models such as GPT-2 and BERT.

\begin{table}[t] \setlength{\tabcolsep}{7pt}
\begin{small}
\centering
\resizebox{0.8\linewidth}{!}{
\begin{tabular}{lccc}
\toprule
Dataset &  & Human & BERT  \\
\midrule 
\multirow{2}{*}{\shortstack{Yelp\\(English)}} & clean & $0.9562\pm 0.0006$ & $0.706$ \\
\cmidrule{2-4}
& adversarial & $0.9390\pm 0.0010$ & $0.000$ \\
\midrule
\midrule
\multirow{2}{*}{\shortstack{THUNews\\(Chinese)}} & clean & $0.9400\pm 0.0014$ & $0.818$ \\
\cmidrule{2-4}
& adversarial & $0.9369\pm 0.0015$ & $0.000$ \\
\bottomrule
\end{tabular}
}
\caption{Human performance compared to BERT classifiers on the original and adversarial datasets. 
}
\label{tab:human}
\end{small}
\end{table}

\paragraph{Utility Preservation Evaluation}
To evaluate human performance on our generated adversarial data, we randomly sample $50$ clean sentences and $50$ adversarial sentences generated by the targeted \attack(+all) for both the English Yelp and the Chinese THUNews dataset. 
For each sentence, we present the annotators with two labels: a ground truth label and a targeted wrong label (e.g., the most opposite sentiment), and request annotators to choose the correct one. 
Both clean text and adversarial text are randomly shuffled. 

The detailed evaluation results with standard deviation are shown in Table \ref{tab:human}. 
We find that our adversarial text barely impacts human perception, as the human performance on adversarial Yelp data is $93.9\%$, only $2\%$ lower than the clean data. Human performance on the adversarial Chinese THUNews is $93.7\%$, which is very close to the performance of $94.0\%$ on the clean dataset.

\section{Related Work}
\label{sec:related}

Our proposed semantic perturbation functions generalize the existing textual adversarial attacks. 

For typo-based perturbation function $\mathcal{F}_T$, existing work \citep{Li2018TextBuggerGA, ebrahimi2017hotflip} applies character-level perturbation to carefully crafted typo words (\emph{e.g.}, from ``foolish'' to ``fo0lish''), thus making the model ignore or misunderstand the original statistical cues. 

Knowledge-based perturbation function $\mathcal{F}_K$ uses knowledge base to constrain the search space. For example, \citet{DBLP:conf/acl/ZangQYLZLS20} uses sememe-based knowledge base from HowNet~\citep{hownet} to construct a search space for word substitution. 

Different from our contextualized semantic perturbation function $\mathcal{F}_C$, other work~\citep{Jin2019IsBR, Li2018TextBuggerGA} uses a non-contextualized word embedding from GLoVe~\citep{DBLP:conf/emnlp/PenningtonSM14} or Word2Vec~\citep{DBLP:conf/nips/MikolovSCCD13} to build synonym candidates, by querying the cosine similarity or euclidean distance between the original and candidate word and selecting the closet ones as the replacements. However, some antonyms also have high cosine similarity in the Word2Vec space. Thus, additional hand-crafted filtering rules are needed to ensure that the meaning is not changed.

Other work \citep{Garg2020BAEBA,li2020bert,li2021contextualized} also leverages pre-trained models to generate contextualized perturbations by masked language modeling, which is a parallel work to \attack, where we explore the BERT embedding clusters to generate high-quality adversarial examples.

In terms of optimization, unlike the \textit{heuristic-based} previous work that uses greedy \citep{Jin2019IsBR} or genetic algorithms \citep{DBLP:conf/acl/ZangQYLZLS20} which search for the optimal perturbations, or \textit{gradient-based} methods \citep{wang-etal-2020-t3,guo2021gradient} which search for perturbation on a tree-autoencoder with only syntactic constraints or  a distribution of adversarial examples, we use an \textit{optimization-based} method to efficiently and effectively search for the optimal adversarial perturbation in the semantic preserving spaces to ensure the validity and naturalness of perturbed sentences.

\section{Conclusion}
In this paper, we propose a novel semantic adversarial attack framework \attack to probe the robustness of LMs. 
Comprehensive experiments show that \m{\attack is able to generate natural adversarial texts in different languages and achieve higher attack success rates than existing textual attacks.
We also demonstrate that existing SOTA LMs and defense methods are still vulnerable to \attack. }
We expect our study to shed light on future research on evaluating and enhancing the robustness of LMs for different languages. 

\subsubsection*{Acknowledgments}
We gratefully thank the anonymous reviewers and meta-reviewers for their constructive feedback. This work is partially supported by the NSF grant No.1910100, NSF CNS 20-46726 CAR, and Sloan Fellowship.

\bibliography{acl}

\clearpage
\appendix
\label{sec:appendix}

\section{Broader Impact}

In this paper, we propose an effective and novel adversarial attack framework \attack to probe the robustness of state-of-the-art NLP models. Our experiments show that even pre-trained large-scale language models for different languages are not robust under \attack. We will open-source our code to shed light on future research to evaluate and enhance the robustness of NLP models. Considering attackers may leverage our code to perform adversarial attacks to NLP models, we suggest using adversarial training as an effective approach to improving adversarial robustness, and our proposed framework has provided an efficient way to generate these adversarial training data.

\section{Model Settings}
\paragraph{Whitebox Classifier}
\label{appdix:model}
For English dataset, we use BERT and self-attention LSTM as the classifiers. BERT is a transformer \cite{Vaswani2017AttentionIA} based model, which is unsupervised pretrained on large corpora. We use the $12$-layer BERT-base model with $768$ hidden units, $12$ self-attention heads, and $110$M parameters. 
For self-attention LSTM, we set the self-attention LSTM to $10$ attention hops internally, and use a $300$-dim BiLSTM and a $512$-units fully-connected layer before the output layer. 

We fine-tune BERT on Yelp dataset with a batch size of $64$, learning rate of $2e-5$ and early stopping. We train the Self-attention LSTM-based model on $500$K review training set for $29$ epochs with stochastic gradient descent optimizer under the initial learning rate of $0.1$.
We run our experiments on i7-7820X CPU with $128$GB memory on one RTX 2080Ti GPU.

For both Chinese datasets, we use BERT \citep{Devlin2019BERTPO} as the classifier. Chinese BERT is a transformer \cite{Vaswani2017AttentionIA} based model, which is unsupervisedly pretrained on large Chinese corpora and is effective for downstream Chinese NLP tasks. We use the 12-layer BERT-base model with 768 hidden units, 12 self-attention heads and 110M parameters. We fine-tune BERT on each dataset independently with a batch size of 64, learning rate of 2e-5 and early stopping.

\paragraph{Blackbox Classifier}
The blackbox LSTM and BERT classifiers are trained/finetuned from scratch. The parameters of blackbox models are different from the whitebox ones.

\section{Dataset Details}
\label{appendix:dataset}
\noindent $\bullet \,$ \textbf{Yelp Dataset} consists of $2.7$M yelp reviews and each one has its corresponding star level to be predicted by our model. The target stars level is an integer number in the inclusive range of $[0, 4]$, which can be treated as $5$ classes. We follow the process in \citet{lin2017structured} to randomly select 500K review-star pairs as the training set, $2,000$ as the development set, and $2,000$ as the test set. 

\noindent $\bullet \,$ \textbf{SNLI Dataset} \cite{snliemnlp2015} consists of $570$k human-written English sentence pairs and each pair contains one premise and one hypothesis. These pairs are manually labeled as entailment, contradiction, or neutral, which can be predicted by our model. We use $550$k pairs as training set, $10$k as the development set, and $10$k as the test set. We follow the baseline setting \citep{li2020bert} and only allow perturbations on hypotheses (Table \ref{tab:english}) or premises (Appendix Table \ref{tab:english_snli_whitebox_p} \& \ref{tab:english_snli_blackbox_p}).

\noindent $\bullet \;$ \textbf{THUNews \citep{sun2016thuctc}} is a public Chinese 14-category news classification dataset. It consists of more than $740$k news articles from Sina News between 2005 and 2011. These articles are classified into $14$ categories, such as education, technology, society and politics. To speed up the evaluation process, we use the news titles for classification. We evenly sample articles from all classes, and use $585,390$ articles as the training set, $250,682$ as the development set, and another $1,000$ as the testing set for the adversarial evaluation. 

\noindent $\bullet \;$ \m{\textbf{Wechat Finance Dataset} is a private dataset from the Wechat team, who collect $13,051$ subscription accounts in the finance domain. They use crowd-sourcing to classify the account into $11$ sub-classes, such as insurance, banks and funds. Each account description has $94.18$ Chinese characters on average. We split the dataset into training set ($10,000$ descriptions), validation set ($1,163$ descriptions) and test set ($1,888$ descriptions).}

\begin{table}[htp!]\small \setlength{\tabcolsep}{7pt}
\centering
\begin{tabular}{cccc}
\toprule
Dataset & avg length &  LSTM Acc & BERT Acc \\
\midrule
Yelp & 135 & 0.705 & 0.706\\
\midrule
SNLI & 13(P)/7(H) & 0.716 & 0.829\\
\bottomrule
\end{tabular}
\caption{Statistics of Yelp Dataset and SNLI Dataset together with benign accuracy of two models. In SNLI Dataset, we calculate the average length of premises (P) and hypotheses (H) separately.}
\label{tab:english_stats}
\end{table}

\begin{algorithm}[t]
\begin{small}
  \caption{\small \attack: Generating multilingual natural adversarial examples} \label{algo}
  \begin{flushleft} 
  \textbf{Input:} Input tokens $\boldsymbol{x} = [\boldsymbol{x_0, x_1, ..., x_n}]$, classifier $f:\boldsymbol{x} \rightarrow \boldsymbol{z}$ maps input to logits, attack objective function $g(\cdot)$,  embedding matrix $\boldsymbol{M_e}$, constants $c$ and $\kappa$, max iteration steps $m$, semantic perturbation function $\mathcal{F}$ \\
  \textbf{Output:} Adversarial text $x'$
  \end{flushleft} 
  \begin{algorithmic}[1]
  \State Initialize perturbation $\boldsymbol{e^*_{{}_{0}}} \leftarrow 0$
  \State $\boldsymbol{e} \leftarrow \boldsymbol{M_e} \boldsymbol{x}$
  \State $\boldsymbol{e'} \leftarrow \boldsymbol{e} + \boldsymbol{e^*_{{}_{0}}}$
  \State $\boldsymbol{x'} \leftarrow \boldsymbol{x}$
  \For{$k=0,1,...,m-1$}
    \State \emph{// Phase I: Optimize over the $\boldsymbol{e^*_{{}_{k}}}$}
    \State$ \mathcal{L}(\boldsymbol{e^*_{{}_{k}}}) \leftarrow  ||\boldsymbol{e^*_{{}_{k}}}||_p + c \cdot g(\boldsymbol{x'})$
    \State $\boldsymbol{e^*_{{}_{k+1}}} \leftarrow \boldsymbol{e^*_{{}_{k}}} - \alpha \nabla \mathcal{L}(\boldsymbol{e^*_{{}_{k}}})$
    \State \emph{// Phase II: Token Substitution}
    \State $\boldsymbol{e'} \leftarrow \boldsymbol{e} + \boldsymbol{e^*_{{}_{k+1}}}$
    \For{$i = 1,2,...,n$} 
        \State $\mathcal{S} = \mathcal{F}(x_i) $  \emph{// Get the perturbation search space}
        \State $ \boldsymbol{x'_i} \leftarrow \argmin_{\boldsymbol{x'_i} \in \mathcal{S}} (  || \boldsymbol{e’_i} - \boldsymbol{M_e x’_i} ||_p)$
    \EndFor
  \EndFor
  \State \textbf{return} $\boldsymbol{x'}$
  \end{algorithmic}
  \end{small}
\end{algorithm}

\section{Experimental Setting}
\label{sec:exp_set}
\subsection{Attack Setup}
\attack is a whitebox attack method which requires access to the model parameters and gradients. However, it can be easily adapted to 
\m{blackbox settings. In our experiment, we consider the following two blackbox settings: a soft-label blackbox setting and a more rigorous zero-query blackbox setting. In soft-label blackbox setting, attackers can only query the classifier for output probabilities on a given input. We adapt our method to this setting by distillation. The output confidence of the blackbox (teacher) model is used to train a student model. Then we run whitebox attacks on the student model and attack the teacher model with adversarial instances provided by the student model. In zero-query blackbox setting, the target models (usually state-of-the-art large-scale language models enhanced with cutting-edge defense methods) are unavailable during the attacking phase, which is a common scenario in real-world applications and better demonstrates the algorithm's ability to generalize across models. We adapt \attack and baseline methods to this setting by performing a transferability-based backbox attack, in which we use adversarial texts created by BERT to attack the target models.}

\subsection{Embedding Space Construction}
To construct the contextualized semantic perturbation function $\mathcal{F}_C$,  we select $22,271$ English words commonly used as $\mathcal{X}$, which is also the vocabulary used by English BERT. For each word, We select at most $100$ sentences that contain this specific word from wikidump. These contextualized embeddings form an embedding space of $2,181,622$ vectors in total. We choose $k = 700$ and $\epsilon = 8$, which means we only choose words that appear more than $8$ times in the $700$ nearest neighbors as the perturbation set $\mathcal{S}$. We apply similar strategies when constructing Chinese BERT embedding space, by choosing $5,178$ Chinese tokens appearing in the training data and up to $100$ sentences from Chinese Wikipedia, which form an embedding space of $508,619$ vectors in total. When performing KNN, we choose $k = 700$ and $\epsilon = 5$. The query time of $\mathcal{F}_C$ is around $2.6$s for English and $0.9$s for Chinese. We provide more detailed settings in Appendix \ref{sec:algo_set}.

\subsection{Semantic Perturbation Functions} \label{appendix:func}
\paragraph{English} We evaluate the following semantic perturbation functions for English corpus: typo-based perturbation function $\mathcal{F}_T$, knowledge-based perturbation function $\mathcal{F}_K$, and contextualized semantic perturbation function $\mathcal{F}_C$ based on BERT embedding clusters, together with the combination of $\mathcal{F}_T$, $\mathcal{F}_K$ and $\mathcal{F}_C$. The average sizes of search spaces obtained by $\mathcal{F}_T$, $\mathcal{F}_K$ and $\mathcal{F}_C$ are $5.03$, $2.38$ and $4.46$, respectively. 

\paragraph{Chinese} We implement semantic perturbation functions for Chinese corpora as follows: ($1$) typo-based perturbation function $\mathcal{F}_T$, where typos are defined as Chinese characters with similar strokes or pronunciations, ($2$) knowledge-based perturbation function $\mathcal{F}_K$, where synonyms are obtained from Chinese WordNet, ($3$) contextualized semantic perturbation function $\mathcal{F}_C$ by Chinese BERT embedding clusters, and ($4$) the combination of these three functions. 

Because in Chinese there are many characters with the same pronunciation, we limit the number of characters obtained by similar pronunciations to $5$. 
The average sizes of perturbation search space collected by $\mathcal{F}_T$, $\mathcal{F}_K$ and $\mathcal{F}_C$ are $8.53$, $0.27$ and $17.06$. $\mathcal{F}_K$ gives fewer candidate perturbations because Chinese WordNet has limited hand-crafted knowledge, while $\mathcal{F}_C$ gives more choices because it searches in BERT embedding space without human supervision. 

\subsection{Attack Hyper-parameter Settings}
For English dataset, we set the max optimization steps $m$ to $100$ and use $\ell_2$ norm in the loss function (equation \ref{cw}) that is iteratively optimized via Adam \citep{Kingma2014AdamAM}. Constants $c$ and $\kappa$ are set to $1e2$ and $1$ in Yelp dataset, $1e4$ and $0$ in SNLI dataset, which result in higher attack success rate and lower perturbation rate based on a series of ablation studies provided in Appendix Figure \ref{fig:hyper}. We set our random seed to $1111$ for reproducibility. 

For Chinese dataset, we follow the experiment setting in English attacks for optimizing adversarial examples and training BERT models. Constants $c$ and $\kappa$ are set to $100$ and $1$ respectively to get the best performance. We set our random seed to $1111$ for reproducibility. We experiment with different attack strategies in Appendix Table \ref{tab:l1norm1} to \ref{tab:strategy}.

\section{\attack Implementation Details}
\label{sec:algo_set}

\subsection{Typo-based Perturbation Function Implementation}

We use the similar Chinese character list\footnote{Publicly available at \url{https://github.com/zzboy/chinese/}} that contains more than 9,000 common Chinese characters. We use the existing Python library\footnote{Publicly available at \url{https://github.com/mozillazg/python-pinyin}} to query the pronunciations for Chinese characters and another library\footnote{Publicly available at \url{https://github.com/letiantian/Pinyin2Hanzi}} to search for the words that share the same pronunciations. Because in Chinese there are many characters with the same pronunciation, we limit the number of characters obtained by similar pronunciations to $5$. 

\subsection{Knowledge-based Perturbation Function Implementation}
In this paper, we use WordNet as an example to illustrate how our framework can integrate the rule-based knowledge to enhance the quality of our adversarial examples. For an input token $x$, we first query the synonym set $s$ in the WordNet. For each meaning of the input word, the output synonym set $s$ contains several synonyms that have this specific meaning. The output synonyms are given with their corresponding part-of-speech tags. In order not to include synonyms that have unusual part of speech, which may result in strange grammatical errors after replacement, we counted the frequency of each part of speech in set $s$ and only selected the words with the highest frequency of part of speech. Using the synonym set after filtering, we are able to generate adversarial input texts that mislead models' prediction while having little effect on human understanding.

\begin{figure}[t]
    \begin{minipage}{0.49\linewidth}
    \centering
    \includegraphics[width=\linewidth]{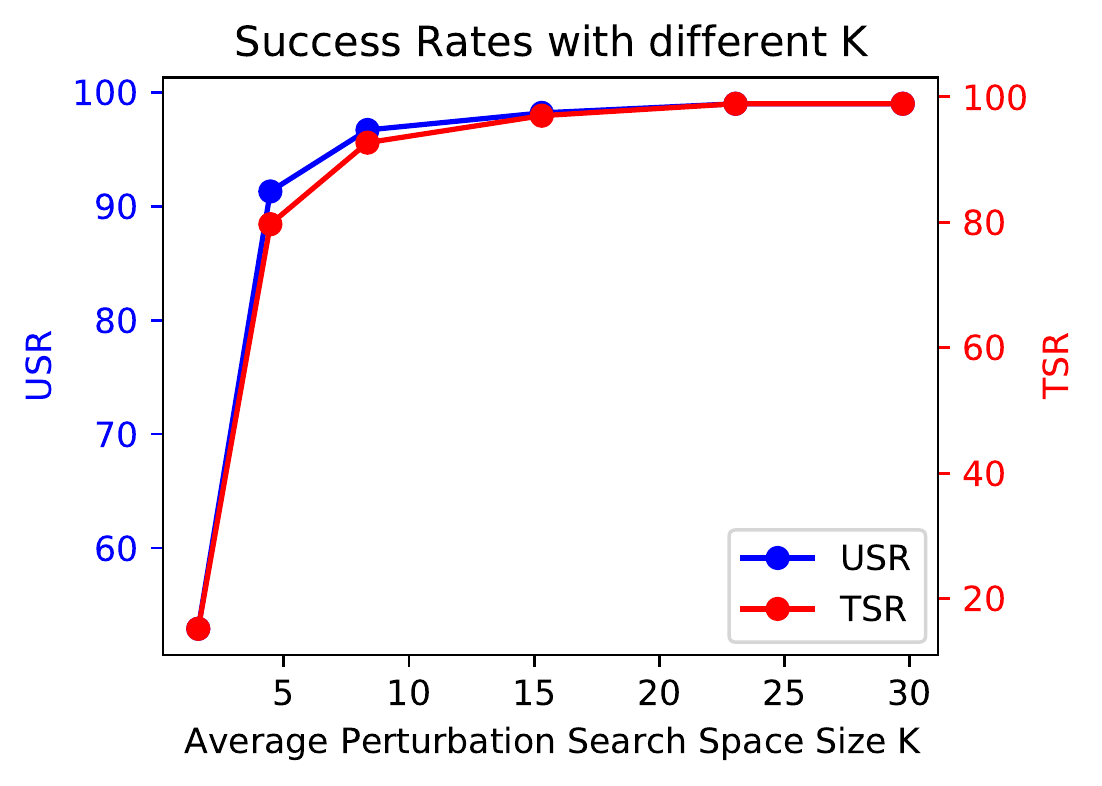}
    \subcaption{Attack success rates with different perturbation search space size K.}
    \label{subfig:KE-sr}
    \end{minipage}
    \begin{minipage}{0.49\linewidth}
    \centering
    \includegraphics[width=\linewidth]{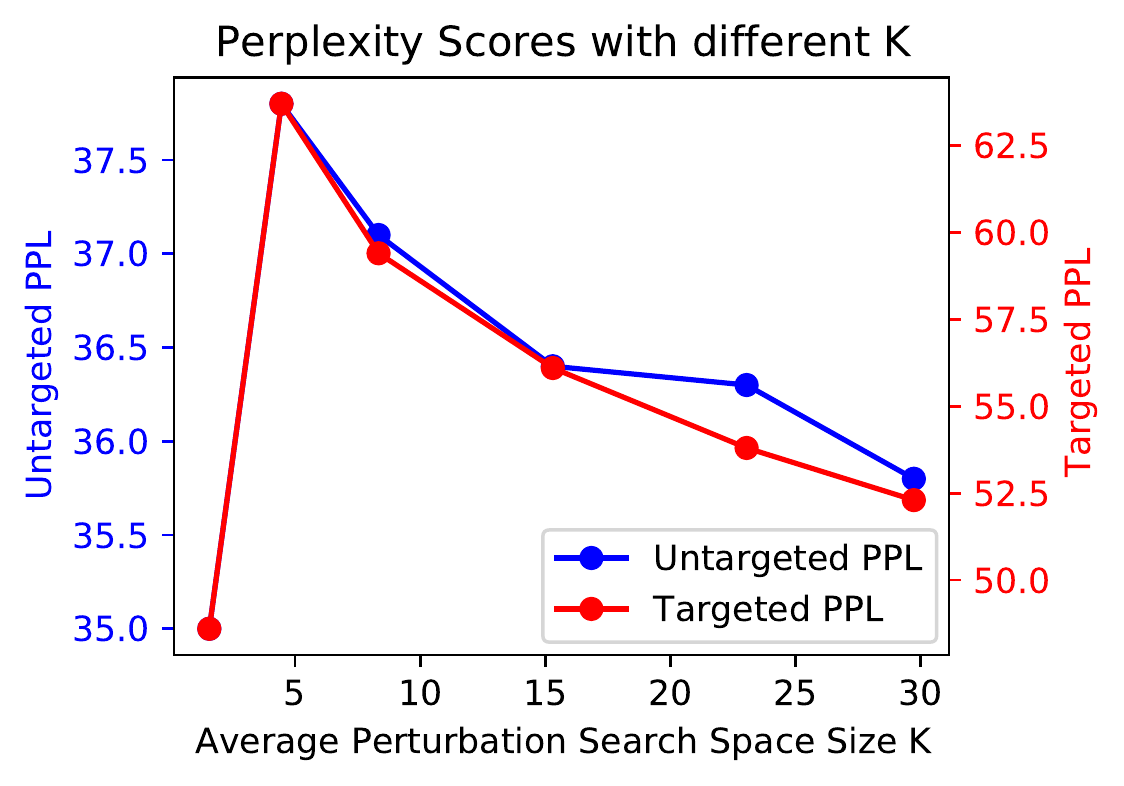}
    \subcaption{Perplexity scores with different perturbation search space size K.}
    \label{subfig:KE-ppl}
    \end{minipage}
    \caption{English perturbation space size selection.}
    \label{fig:KE}
\end{figure}

\begin{figure}[t]
    \begin{minipage}{0.49\linewidth}
    \centering
    \includegraphics[width=\linewidth]{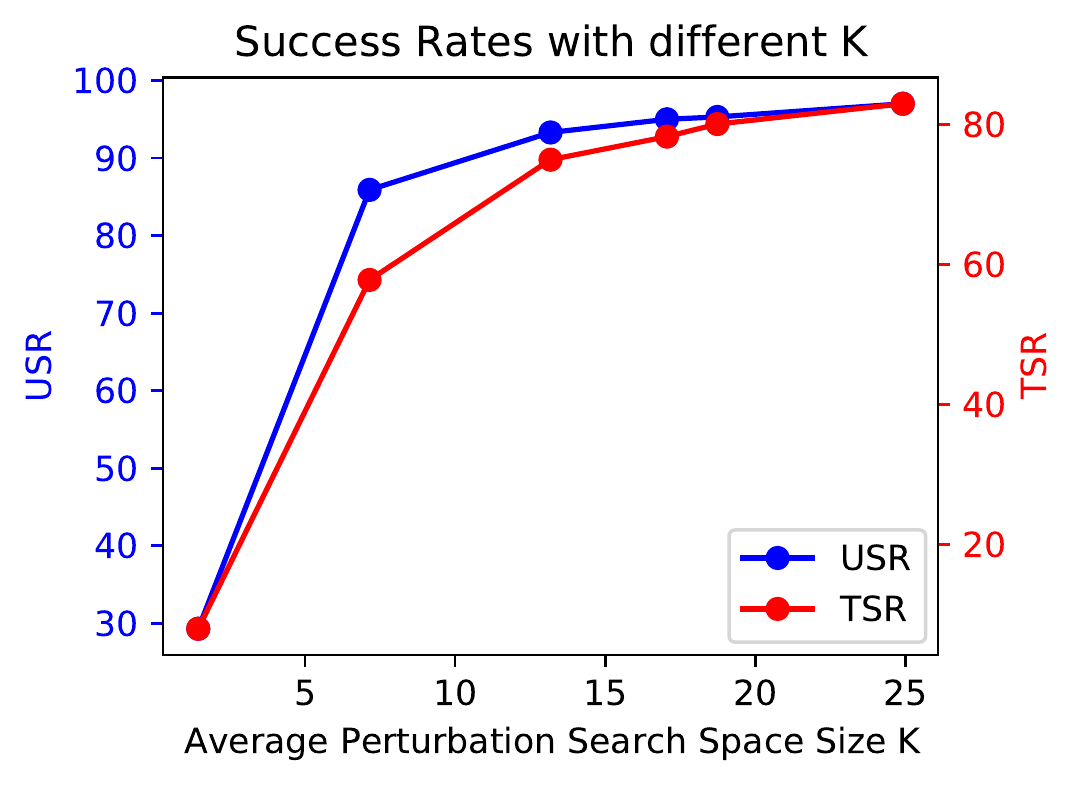}
    \subcaption{Attack success rates with different perturbation search space size K.}
    \label{subfig:KC-sr}
    \end{minipage}
    \begin{minipage}{0.49\linewidth}
    \centering
    \includegraphics[width=\linewidth]{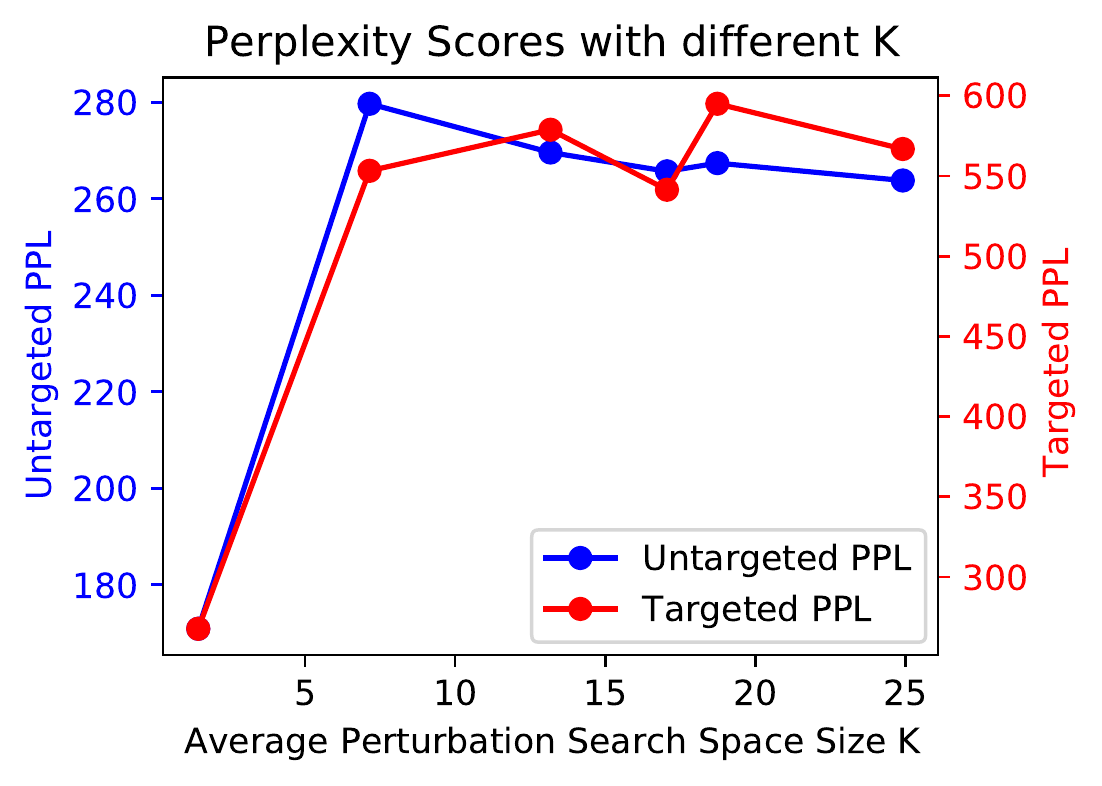}
    \subcaption{Perplexity scores with different perturbation search space size K.}
    \label{subfig:KC-ppl}
    \end{minipage}
    \caption{Chinese perturbation space size selection.}
    \label{fig:KC}
\end{figure}

\begin{table*}[t]\small \setlength{\tabcolsep}{7pt}
\centering
\begin{subtable}[t]{.48\textwidth}
\centering
\resizebox{1.0\linewidth}{!}{
\begin{tabular}{ccccc}
\toprule
Model & Method & \% USR/TSR  & \% Perturbation\\
\midrule
\multirow{6}{*}{\shortstack{BERT\\(Acc: \\0.706)}}
& TextFooler   & 84.7/48.6 & 13.5/32.2\\
& BERT-ATTACK   & \textbf{95.4}/71.1 & 4.2/11.2\\
& \attack (+$\mathcal{F}_T$)   & 32.6/6.7 & 4.6/9.1\\
& \attack (+$\mathcal{F}_K$)   & 58.8/51.5 & 5.9/15.5\\
& \attack (+$\mathcal{F}_C$)  & 68.4/61.3 & 4.7/12.1\\
& \attack (+all)  & 67.5/\textbf{72.4} & 4.0/11.7\\
\midrule
\multirow{6}{*}{\shortstack{Self-Attention \\LSTM\\(Acc:\\ 0.705)}}
& TextFooler & 17.5/5.7  & 9.6/28.0\\
& BERT-ATTACK   & 65.0/24.7 & 2.2/3.7\\
& \attack (+$\mathcal{F}_T$)  & 51.2/25.0 & 18.3/22.4\\
& \attack (+$\mathcal{F}_K$)  & 39.2/24.0 & 15.0/19.2\\
& \attack (+$\mathcal{F}_C$)  & 57.7/33.7 & 23.4/26.7\\
& \attack (+all)  & \textbf{74.1}/\textbf{67.0} & 30.6/35.8\\
\bottomrule
\end{tabular}
}
\caption{Yelp Dataset}
\label{tab:english_yelp_blackbox}
\end{subtable}
\quad
\begin{subtable}[t]{.48\textwidth}
\centering
\resizebox{1.0\linewidth}{!}{
\begin{tabular}{ccccc}
\toprule
Model & Method & \% USR/TSR  & \% Perturbation\\
\midrule
\multirow{6}{*}{\shortstack{BERT\\(Acc: \\0.829)}}
& TextFooler   & 73.2/30.8 & 22.3/24.7\\
& BERT-ATTACK   & \textbf{88.9}/\textbf{61.8} & 17.0/20.1\\
& \attack (+$\mathcal{F}_T$)   & 19.1/6.8 & 10.2/11.2\\
& \attack (+$\mathcal{F}_K$)   & 36.7/12.5 & 12.9/20.0\\
& \attack (+$\mathcal{F}_C$)  & 59.8/45.0 & 14.8/26.1\\
& \attack (+all)  & 63.9/40.5 & 15.2/17.1\\
\midrule
\multirow{6}{*}{\shortstack{Self-Attention \\LSTM\\(Acc:\\ 0.705)}}
& TextFooler   & 52.9/24.2 & 20.1/24.7\\
& BERT-ATTACK   & 62.8/36.9 & 17.9/18.7\\
& \attack (+$\mathcal{F}_T$)   & 49.9/33.3 & 26.4/32.9\\
& \attack (+$\mathcal{F}_K$)   & 40.3/22.5 & 22.1/25.6\\
& \attack (+$\mathcal{F}_C$)  & 68.9/56.9 & 33.0/39.5\\
& \attack (+all)  & \textbf{75.4}/\textbf{57.0} & 42.3/37.9\\
\bottomrule
\end{tabular}
}
\caption{SNLI Dataset}
\label{tab:english_snli_blackbox}
\end{subtable}
\caption{
\small \m{Soft-label blackbox attack success rate for different attacks under targeted/untargeted attacks (TSR/USR) and corresponding word perturbation percentage against self-attention LSTM and BERT on Yelp and SNLI datasets.}} 
\label{tab:english_blackbox}
\end{table*}

\section{Ablation Studies}
\label{appdix:analysis}

\subsection{Perturbation space size selection}
In Figure \ref{fig:KE}, \ref{fig:KC}, we present the attack success rates and perplexity scores of generated adversarial examples under different sizes of perturbation search space. We observe that in both languages, larger K lead to higher attack success rates. In English, PPL score decreases when K continues to increase, while in Chinese PPL score remains at a similar level.

\begin{figure}[t]
    \begin{minipage}{0.49\linewidth}
    \centering
    \includegraphics[width=\linewidth]{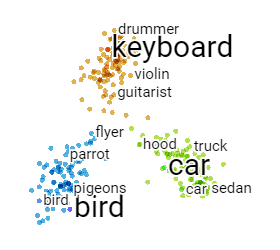}
    \subcaption{Visualization. }
    \label{fig:bert_clusters}
    \end{minipage}
    \begin{minipage}{0.49\linewidth}
    \centering
    \includegraphics[width=\linewidth]{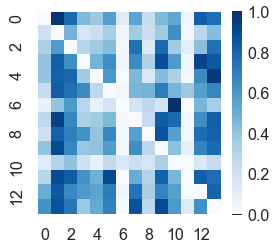}
    \subcaption{Confusion matrix.}
    \label{fig:ch_conf_matrix}
    \end{minipage}
    \caption{Ablation studies. (a) shows the visualization of English words in BERT embedding clusters. (b) shows the TSR confusion matrix on THUNews dataset.}
    \label{fig:ablation_study}
\end{figure}

\subsection{Attack Efficiency}
\attack is more efficient than existing baselines since it can substantially decrease the query time when performing attacks. SemAttack searches for the optimal perturbation $e^*$ for a whole sentence in one query, instead of querying every word. Quantitatively, \attack is designed to query the model for less than 100 iterations, while BERT-ATTACK and TextFooler require hundreds of queries to generate one adversarial example on average.

\subsection{BERT Embedding Space}
\label{appendix:embedding}
In Figure \ref{fig:bert_clusters}, we visualize three clusters: ``car'', ``bird'' and ``keyboard''. Here ``keyboard'' is used as an instrument, not a peripheral device of PCs. As we can see, `bird' has neighbors such as ``pigeons'', ``parrot'' and ``flyer'', which are not present in knowledge space. Word ``keyboard'' has neighbors such as ``drummer'', ``violin'' and ``guitarist'', which are contextualized based on the query context.

\subsection{Additional Results on Attacking SNLI}

We follow the setting of \citep{li2020bert} and perturb only hypotheses or premises for SNLI tasks. Attack results for perturbing hypotheses are shown in main paper Table \ref{tab:english}. Attack results for perturbing premises only are shown in Table \ref{tab:english_snli_whitebox_p} and \ref{tab:english_snli_blackbox_p}.

\begin{table}[t]\small \setlength{\tabcolsep}{7pt}
\centering
\resizebox{1.0\linewidth}{!}{
\begin{tabular}{cccc}
\toprule
Model & Method & \% USR/TSR  & \% Perturbed\\
\midrule
\multirow{5}{*}{\shortstack{BERT\\(Acc: 0.829)}} & HotFlip   & 43.6/20.5 & 28.0/29.8\\
& \attack (+$\mathcal{F}_T$)   & 11.6/4.1 & 11.2/12.5\\
& \attack (+$\mathcal{F}_K$)   & 25.4/12.2 & 12.9/17.2\\
& \attack (+$\mathcal{F}_C$)  & 66.4/36.7 & 16.4/21.2\\
& \attack (+all)  & \textbf{72.7}/\textbf{46.1} & 17.5/21.6\\
\midrule
\midrule
\multirow{5}{*}{\shortstack{Self-Attention \\LSTM\\(Acc: 0.716)}} & HotFlip   & 10.8/8.2 & 10.2/10.0\\
& \attack (+$\mathcal{F}_T$)   & 47.5/29.3 & 15.5/19.1\\
& \attack (+$\mathcal{F}_K$)   & 43.4/22.2 & 13.2/15.0\\
& \attack (+$\mathcal{F}_C$)  & 69.7/48.5 & 28.2/35.5\\
& \attack (+all)  & \textbf{70.7}/\textbf{46.5} & 29.5/36.6\\
\bottomrule
\end{tabular}
}
\caption{
The whitebox attack success rate (in terms of ``USR/TSR'') and corresponding word perturbation percentage against LSTM and BERT on the SNLI dataset by only perturbing premises.} 
\label{tab:english_snli_whitebox_p}
\vspace{-5mm}
\end{table}

\begin{table}[t]\small \setlength{\tabcolsep}{7pt}
\centering
\resizebox{1.0\linewidth}{!}{
\begin{tabular}{cccc}
\toprule
Model & Method & \% USR/TSR  & \% Perturbed\\
\midrule
\multirow{6}{*}{\shortstack{BERT\\(Acc: 0.829)}} & TextFooler   & 61.3/31.1 & 15.0/17.0\\
& BERT-ATTACK   & \textbf{60.2}/\textbf{34.8} & 25.6/34.4\\
& \attack (+$\mathcal{F}_T$)   & 11.5/4.3 & 4.9/5.6\\
& \attack (+$\mathcal{F}_K$)   & 17.0/7.0 & 11.2/13.1\\
& \attack (+$\mathcal{F}_C$)  & 43.0/24.8 & 13.4/16.1\\
& \attack (+all)  & 47.0/30.2 & 14.6/17.5\\
\midrule
\midrule
\multirow{6}{*}{\shortstack{Self-Attention \\LSTM\\(Acc: 0.716)}} & TextFooler   & 19.1/10.6 & 10.3/10.6\\
& BERT-ATTACK   & {42.9}/{31.5} & 19.4/23.0\\
& \attack (+$\mathcal{F}_T$)   & 29.4/22.7 & 23.1/27.6\\
& \attack (+$\mathcal{F}_K$)   & 23.2/15.8 & 20.7/23.0\\
& \attack (+$\mathcal{F}_C$)  & 55.9/46.3 & 43.5/45.7\\
& \attack (+all)  & \textbf{59.0/49.7} & 45.7/47.8\\
\bottomrule
\end{tabular}
}
\caption{
The blackbox attack success rate (in terms of ``USR/TSR'') and corresponding word perturbation percentage against LSTM and BERT on the SNLI dataset by only perturbing premises.} 
\label{tab:english_snli_blackbox_p}
\vspace{-5mm}
\end{table}

\subsection{Ablation Studies on Attack Capability}
In this section, we will evaluate the possible factors that will affect the attack success rate. Here, we set the candidate search space $\mathcal{S}$ to be the whole vocabulary $V$ to eliminate variables introduced by the perturbation function.

\subsection{Norm selection} 
In the main experiment, we use $l_2$ norm for our attack loss function (equation 7). However, because $l_1$ norm is known for good at feature selection and generating sparse features, we conduct the following experiments by setting $l_p$ to $l_1$ and make an comparison with $l_2$ norm. The experimental results are shown in Table \ref{tab:l1norm1} and \ref{tab:l1norm2}. We find the overall attack success rates decrease when switching to $l_1$ norm. However, given the same set of constants $c$ and $\kappa$, we find the $l_1$ attack does change less words.

\begin{table*}[t]\small \setlength{\tabcolsep}{7pt}
\centering
\begin{tabular}{cccccccccc}
\toprule
\multirow{2}{*}{Dataset} & Original & & \multicolumn{3}{c}{\attack ($l_2$ untargeted)} & \multicolumn{3}{c}{\attack ($l_1$ untargeted)} & Baseline\\
\cmidrule(lr){4-6} \cmidrule(lr){7-9}  \cmidrule(lr){10-10}
& Acc & $c/k$ & $5/5$ & $10/5$  & $10/10$ & $10/10$ & $10/100$  & $20/20$ & (untargeted) \\
\midrule
\multirow{3}{*}{THUCTC} & \multirow{3}{*}{0.818} & target & -  & - & - & - & -  & -  & -    \\
    &  & untarget & \textbf{1.000}          & \textbf{1.000}  & \textbf{1.000}  & 0.983 &  0.983 & \textbf{0.995}  & 0.040\\
    & & \#/chars & \textbf{1.583}         & 1.690   & 1.718 & \textbf{1.577} &  1.614 & 1.884 & 2.000 \\
\bottomrule
\end{tabular}
\caption{Untargeted attack success rates on Chinese BERT-based classifier for THUCTC dataset. ``target'' and ``untarget'' calculate the targeted attack success rate (equation \ref{eq:tsr}) and the untargeted attack success rate (equation \ref{eq:usr}). ``\#/chars'' counts the number characters are modified in average.
}
\label{tab:l1norm1}
\end{table*}

\begin{table*}[t]\small \setlength{\tabcolsep}{7pt}
\centering
\begin{tabular}{cccccccccc}
\toprule
\multirow{2}{*}{Dataset} & Original & & \multicolumn{3}{c}{\attack ($l_1$ targeted)} & \multicolumn{3}{c}{\attack ($l_2$ targeted)} & Baseline\\
\cmidrule(lr){4-6} \cmidrule(lr){7-9}  \cmidrule(lr){10-10}
& Acc & $c/k$ & $10/10$ & $10/20$  & $30/30$ & $5/5$ & $10/5$  & $10/10$ & (untargeted) \\
\midrule
\multirow{3}{*}{THUCTC} & \multirow{3}{*}{0.818} & target & 0.797  & 0.797 &\textbf{ 0.898 }& 0.941 & \textbf{0.945}  & \textbf{0.945}    & -    \\
    &  & untarget & 0.828          & 0.828 & \textbf{0.920}  & 0.953 &  \textbf{0.958} &  \textbf{0.958} & 0.040\\
    & & \#/chars & 2.000 & \textbf{1.956}   & 3.280 & \textbf{2.924} &  3.186 & 3.045 & 2.000 \\
\bottomrule
\end{tabular}
\caption{Targeted attack success rates on Chinese BERT-based classifier for THUCTC dataset. ``target'' and ``untarget'' calculate the targeted attack success rate (equation \ref{eq:tsr}) and the untargeted attack success rate (equation \ref{eq:usr}). ``\#/chars'' counts the number charcters are modified in average. 
}
\label{tab:l1norm2}
\end{table*}

\subsection{Attack Strategy}\label{sec:strategy}
As we have achieved $100\%$ attack success rate in the untargeted attack scenario, we now focus on the targeted attack scenario and see which factor contributes to the targeted attack success rate. It is straightforward to think different targeted attack strategies will impact the targeted attack success rate, because maybe some classes look "farther" than semantic closer classes. So we tried two strategies on THUCTC dataset: 1) as used in the main paper, we set the targeted false class as ``technology news''. 2) we enumerate all the classes and set the targeted false class to be numerically the next class index. The targeted attack success rate is shown in Table \ref{tab:strategy}. We do find choosing different attack strategies will impact the attack success rate.

\begin{table*}[t]\small \setlength{\tabcolsep}{7pt}
\centering
\begin{tabular}{cccccc}
\toprule
\multirow{2}{*}{Dataset} & Original & & \multicolumn{2}{c}{\attack (targeted $c/\kappa=10/10$)} & Baseline\\
\cmidrule(lr){4-5}  \cmidrule(lr){6-6}
& Acc &  &  strategy 1   &  strategy 2 & (untargeted) \\
\midrule
\multirow{3}{*}{THUCTC} & \multirow{3}{*}{0.818} & target &  \textbf{0.945} & 0.903    & -    \\
    &  & untarget  &  \textbf{0.958} & 0.913 & 0.040\\
    & & \#/chars  & 3.045 & 4.543 & 2.000 \\
\bottomrule
\end{tabular}
\caption{Attack success rates on Chinese BERT-based classifier for two datasets. ``target'' and ``untarget'' calculate the targeted attack success rate (equation \ref{eq:tsr}) and the untargeted attack success rate (equation \ref{eq:usr}). ``\#/chars'' counts the number characters are modified in average.
}
\label{tab:strategy}
\end{table*}

\begin{figure}[t]
    \begin{minipage}{0.49\linewidth}
    \centering
    \includegraphics[width=\linewidth]{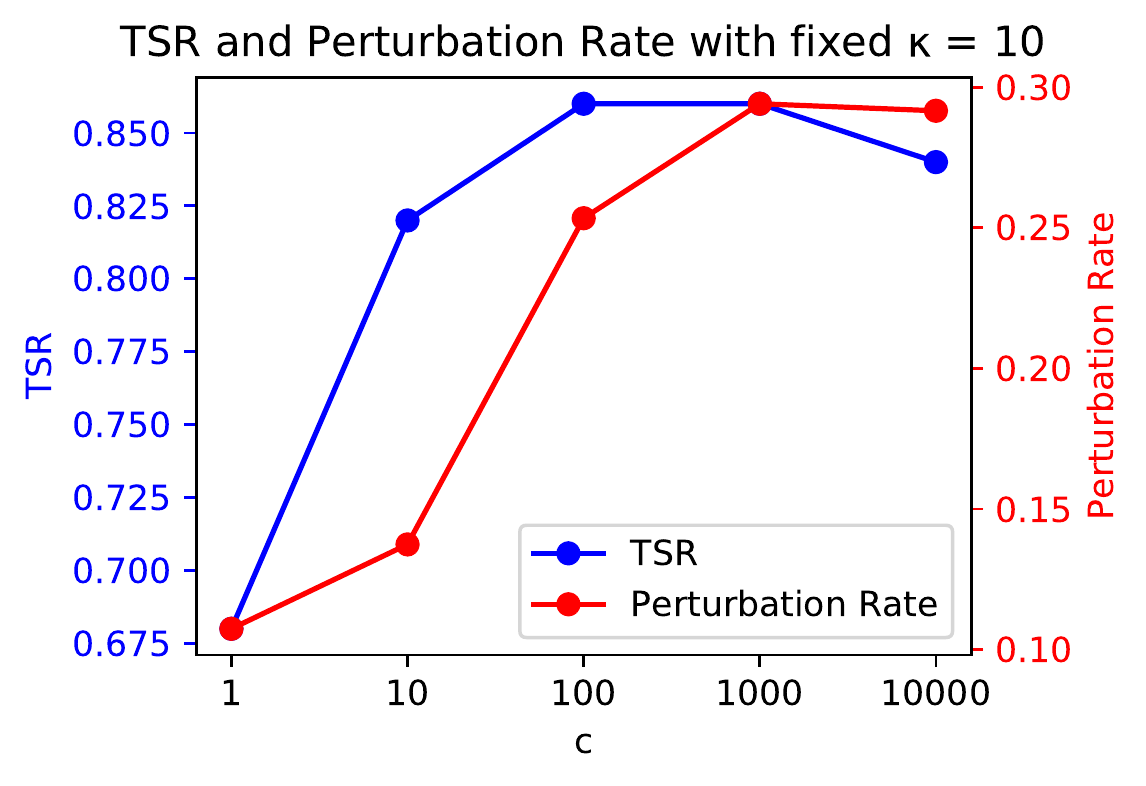}
    \subcaption{Fixed $\kappa$ and different $c$.}
    \label{subfig:hyper-c}
    \end{minipage}
    \begin{minipage}{0.49\linewidth}
    \centering
    \includegraphics[width=\linewidth]{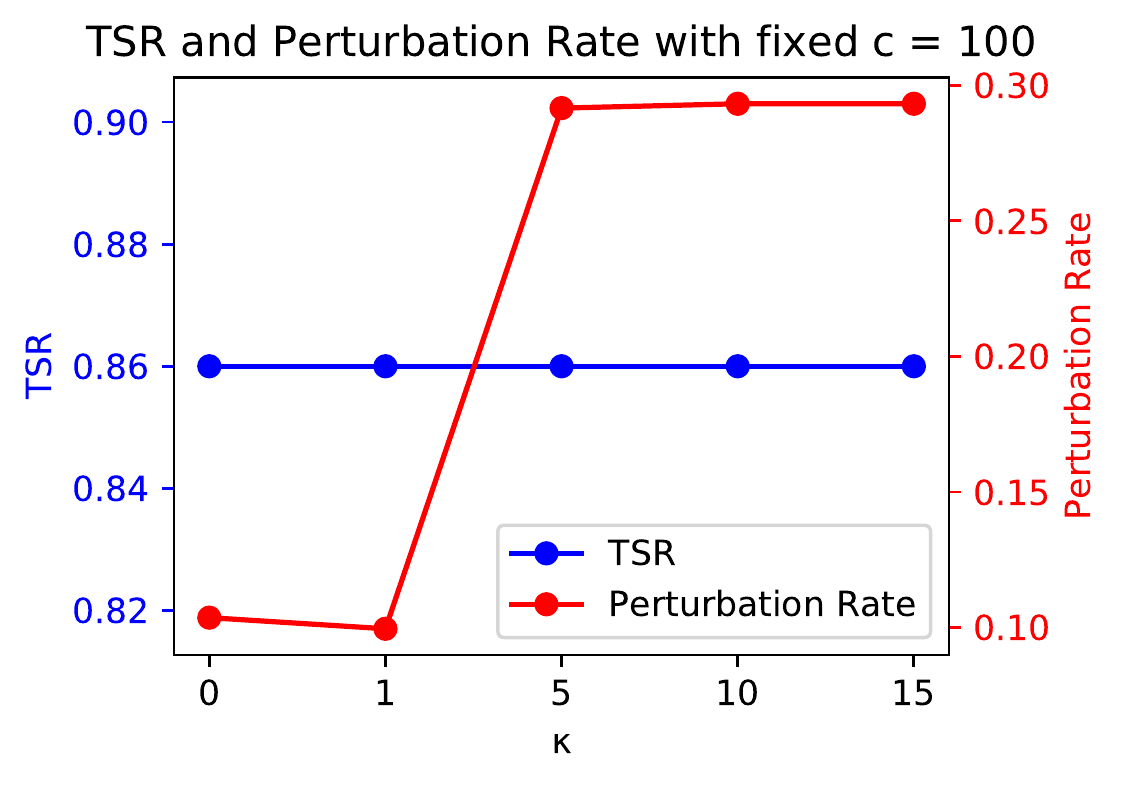}
    \subcaption{Fixed $c$ and different $\kappa$.}
    \label{subfig:hyper-kappa}
    \end{minipage}
    \caption{Hyper parameter selection. In Figure \ref{subfig:hyper-c}, we first fix $\kappa = 10$ and test different $c$ to see how TSR and perturbation rate will change. we test $c = 1, 10, 10^2, 10^3, 10^4$ and find best $c = 100$ to obtain the highest TSR with less perturbations. A smaller or larger $c$ will result in a low TSR or a high perturbation rate. In Figure \ref{subfig:hyper-kappa}, after fixing $c = 100$, we test $\kappa = 0, 1, 5, 10, 15$. We find that $\kappa$ has little influence on TSR while it can change perturbation rate dramatically. A smaller $\kappa$ is able to effectively limit the number of words to be changed. In our experiment, we choose $\kappa = 0, 1$.}
    \label{fig:hyper}
\end{figure}

\subsection{Hyper Parameter Selection}
We have two constants in our attack algorithm, $c$ and $\kappa$, which control the attack success rates and the perturbation rates in our experiments. In order to find out the impact of these hyper parameters, we test with several combinations of different $c$ and $\kappa$. We test on Yelp Dataset and we use BERT as our model. We show our results in Figure \ref{fig:hyper}. As shown in Figure \ref{subfig:hyper-c}, we first fix $\kappa = 10$ and test how TSR and perturbation rate will change according to different $c$. We find that under the same $\kappa$, $c$ mainly controls the attack success rate at the cost of perturbation rate. In some certain range, a larger $c$ encourages the algorithm to achieve our attack goal with the expense of more substitutions. And after exceeding a certain value, TSR will start to decrease while perturbation rate remains high. We then fix $c = 100$ and test different $\kappa$. We show our results in Figure \ref{subfig:hyper-kappa}. We find that $\kappa$ doesn't help to increase TSR and a smaller $\kappa$ helps to limit the words changed without affecting TSR.

For hyper-parameter selection for Chinese datasets, we witness the same phenomenon in English attacks that increasing constant $c$ can improve the attack success rate at the cost of more perturbed characters, while lowering constant $\kappa$ limits the perturbation rate without affecting the attack success rate. 

\subsection{Vulnerability Between Classes}
In THUNews dataset, the article titles are classified into $14$ categories. In order to find out the vulnerability of each class, we test the attack success rate of each source class and target class. The heatmap of results is provided in Figure \ref{fig:ch_conf_matrix}. We find that ``technology news'' and ``entertainment news'' as target classes have higher average success rates than other classes, while ``lottery ticket'' is the lowest. We also find that ``constellation 
news'' has the highest average success rate as source class, while ``sports news'' has the lowest, which means ``constellation news'' is vulnerable and easy to attack while ``sports news'' is much more robust.

\begin{table}[t] \small \setlength{\tabcolsep}{7pt}
\centering
\resizebox{\linewidth}{!}{
\begin{tabular}{cccc}
\toprule
Transfer & Method & \% TSR  & \% USR \\
\midrule
\multirow{6}{*}{\shortstack{Self-Attention \\LSTM \\ $\rightarrow$ \\ BERT}} & TextFooler & 42.4  & 43.9 \\
& BERT-ATTACK  & 8.1 & 33.5\\
& \attack(+$\mathcal{F}_T$)  & 44.4& 32.5\\
& \attack(+$\mathcal{F}_K$)  & 57.7 & 62.0\\
& \attack(+$\mathcal{F}_C$)   & \textbf{74.3} & \textbf{81.2}\\
& \attack(+all)  & 70.0 & 79.8\\
\midrule
\midrule
\multirow{6}{*}{\shortstack{BERT \\ $\rightarrow$ \\ Self-Attention \\LSTM}} & TextFooler   & 30.8 & 31.9\\
& BERT-ATTACK  & 17.6 & 28.5\\
& \attack(+$\mathcal{F}_T$)  & 26.8 & 34.6\\
& \attack(+$\mathcal{F}_K$)   & 35.3 & 35.6\\
& \attack(+$\mathcal{F}_C$)  &\textbf{35.5} & \textbf{36.0} \\
& \attack(+all)  & 30.9 & 31.0\\
\bottomrule
\end{tabular}
}
\caption{Targeted and untargeted attack success rate of transferability attack on Yelp Dataset, evaluating adversarial examples generated against Self-attention LSTMs on BERT, and vice versa.}
\label{tab:english_transfer}
\vspace{-3mm}
\end{table}

\subsection{Transferability Analysis} 
We evaluate the transferability of adversarial examples between different models by attacking a blackbox BERT classifier by using adversarial text generated from a whitebox LSTM, and vice versa.

The transferability-based attack results on Yelp Dataset are shown in Table \ref{tab:english_transfer}. We find that the robustness of the two models is highly different from each other. When we feed adversarial texts generated from the LSTM model into the blackbox BERT model, attack success rate is higher than $70\%$. However, when we test the performance of the blackbox LSTM model on adversarial texts generated from the whitebox BERT, attack success rate is around $30\%$, which is much lower than previous experiment. These results show that Self-Attention LSTMs are more robust than BERT models, and the adversarial examples generated from a robust model has higher attack transferability than non-robust one. Therefore, we can attack blackbox BERT models using a strong Self-Attention LSTM trained by ourselves to generate adversarial texts with high success rates.
We also observe that the USR of transferability-based attack is generally higher than that of targeted attack. Particularly, we achieve the highest success rate of $81.2\%$ when attacking blackbox BERT with text generated by LSTM attacks under untargeted setting.

Furthermore, we find that the adversarial examples generated by the contextualized semantic perturbation functiuon $\mathcal{F}_C$ have the highest attack transferability, which suggests that our contextualized semantic perturbation is more generalizable than rule-based perturbation functions.

\section{Human Evaluation Details}
\label{appendix:human}

\paragraph{Language Quality Evaluation Details} 
We use Amazon Turk for English adversarial example quality annotations, and Alibaba Cloud for Chinese example quality annotations. Each sentence is annotated by 5 annotators. This evaluation only evaluates language quality and grammatical correctness, and thus does not require additional background or domain knowledge. 

We present the annotation instructions on Amazon Turk below.

Please \textbf{rate the language quality} (from 1 to 5, in terms of coherence, fluency, and grammar correctness) of    the presented sentence. 5 means the best language quality, and 1 means the lowest language quality.
 \begin{itemize}
     \item 5: The sentence looks totally correct. There are no grammatical errors. I can fully understand the sentence.
     \item                                 4: The sentence looks somewhat correct. There are one or two grammatical errors or typos. But I can mostly understand the sentence. 
    \item                                 3: The sentence looks OK to me. There are some grammatical errors or typos. I can partly understand the sentence. 
    \item                                2: The sentence looks bad to me. There are grammatical errors or typos everywhere. I can understand it a little.
    \item                                 1: The sentence totally does not make any sense. I cannot understand it.

 \end{itemize}

\paragraph{Utility Preservation Evaluation Details}  We use the targeted \attack to generate the adversarial dataset with  with $c/\kappa=100/1$.
In total, we collected annotations from $21$ graduate students from US universities for English datasets and $26$ annotators from native Chinese speakers for Chinese datasets. Both classification tasks do not require domain knowledge.
The detailed human performance results are shown in Table \ref{tab:human}.

\onecolumn
\section{Perturbation Search Space Examples}

\subsection{English Perturbation Search Space $\mathcal{S}$ Examples}
\begin{table*}[htp!]\small \setlength{\tabcolsep}{7pt}
\centering
\caption{English Perturbation Search Space $\mathcal{S}$ Examples Generated by \attack for BERT-based Classifier using $\mathcal{F}_T$, $\mathcal{F}_K$ and $\mathcal{F}_C$. In the first example, we list some words and corresponding candidate sets generated by these functions. We can see that words generated by $\mathcal{F}_C$ reflect the meaning of the current context. For example, when we say that a hotel is \textit{good}, we may say it's \textit{spacious}. When word \textit{come} is followed by \textit{back}, we may mean \textit{return}. In the following two examples, we show that the same word may have different perturbation sets in different contexts. In the second example, by using \textit{order}, the person means that he ordered food. Considering the context, $\mathcal{F}_C$ provides \textit{eat}, \textit{taste} in its candidate set. In the last example, \textit{order} means the person orders a drink. As a consequence, we have \textit{drink} as a verb with a similar meaning in its candidate set.}
\label{appendix:engperturbationset}
\begin{tabular}{p{13.8cm}}
\toprule

\textbf{Input English Text: } This was my fifth time traveling to vegas! I have stayed at hotels such as the Bellagio, Aria, Cosmopolitan, the venetian, and fortunately enough got a chance to \textbf{stay} at vdara. Considering the reviews I didn't expect vdara to be that-\textbf{good} of a hotel! Vdara was extremely \textbf{clean}, very modern, new, great customer service, \textbf{close} to the strip-connected to the bellagio. easy access to casinos and heart of the strip. Definitely \textbf{coming} back to vegas and booking a room at vdara.
 \\
 \\
$\mathcal{F}_T(stay) = $ stau, stay, etay, stxy, sty \\
$\mathcal{F}_K(stay) = $ quell, last out, bide, persist, stay \\
$\mathcal{F}_C(stay) = $ staying, stay, vacationing, stays, relax, internship, enroll, stayed, visit, settle \\
\\
$\mathcal{F}_T(good) = $ good, gopd, god, gpod, 6ood \\
$\mathcal{F}_K(good) = $ estimable, adept, full, effective, dear, beneficial, dependable, good \\
$\mathcal{F}_C(good) = $ spacious, marvelous, marvel, wonderful, good \\
\\
$\mathcal{F}_T(clean) = $ caeln, coean, clean, cl3an, celan, clen, claen, vlean \\
$\mathcal{F}_K(clean) = $ blank, clean, uninfected \\
$\mathcal{F}_C(clean) = $ spacious, luxurious, lively, vibrant, cleanest, cozy, cleaned, renovated, clean \\
\\
$\mathcal{F}_T(close) = $ close, csole, clode, clsoe, cloe, cloze, c1ose, clse \\
$\mathcal{F}_K(close) = $ close, conclude, close up \\
$\mathcal{F}_C(close) = $ connected, near, close, nearer, closeness \\
\\
$\mathcal{F}_T(coming) = $ c0ming, coimng, conimg, comng, coming, comjng, comlng, cmoing, cming \\
$\mathcal{F}_K(coming) = $ come, derive, issue forth, arrive, hail, total, occur, do, fall \\
$\mathcal{F}_C(coming) = $ returning, traveling, transferring, staying, relocating, visiting, talking, coming \\

\midrule

\textbf{Input English Text: } Stopped by this place for lunch . \textbf{Ordered} the veggie slice and patty they put lettuce cheese and mayo in it and both the slice and patty were amazing. Definitely will be back for more. \\
\\
$\mathcal{F}_T(Ordered) = $ ordered, orered, ordeged, ordeed, orderfd, lrdered, orreded \\
$\mathcal{F}_K(Ordered) = $ rate, ordain, arrange, order, regulate \\
$\mathcal{F}_C(Ordered) = $ ate, tasted, ordered \\

\midrule

\textbf{Input English Text: } Love this speakeasy bar. Last time I was at this location it was still the Panda bar. The place itself is super cozy and intimate. We went there to grab a drink before our Ali Wong show. Hubby \textbf{ordered} a Hendricks gin tonic (12\$-happy hour price?) and I got a cocktail with Pimms (9\$ before 9pm). The drinks were HUMONGOUS! So much so I couldnt finish mine and hubby was tipsy lol. \\
\\
$\mathcal{F}_T(Ordered) = $ ordered, ordeerd, ordred, ordeed, orderfd, oedrred, orreded, ordersd, orderex \\
$\mathcal{F}_K(Ordered) = $ rate, ordain, arrange, order, regulate \\
$\mathcal{F}_C(Ordered) = $ ate, drank, ordered \\

\bottomrule
\end{tabular}
\end{table*}

\newpage 

\begin{CJK*}{UTF8}{gbsn}
\subsection{Chinese Perturbation Search Space $\mathcal{S}$ Examples}
\label{appendix:perturbationset}
\begin{table*}[htp!]\small \setlength{\tabcolsep}{7pt}
\centering
\caption{Chinese Perturbation Search Space $\mathcal{S}$ Examples Generated by \attack for BERT-based Classifier using $\mathcal{F}_T$, $\mathcal{F}_K$ and $\mathcal{F}_C$. Chinese characters are intrinsically polysemous, which requires candidate characters to be contextualized. We list four examples here. In the first two examples, we show two different meanings of character ``美'' in two different sentences. One referring to \textit{the US} which has some other countries' names in its perturbation set, another meaning \textit{poignant} which is used as an adjective. In the last two examples, we show ``长'', a well-known Chinese character that has multiple pronunciations and multiple meanings. We show that our two perturbation functions return different candidate sets. In the third example, ``长'' means a job title, while in the last example it means \textit{growth}.}
\begin{tabular}{p{13.8cm}}
\toprule

\textbf{Input Chinese Text: } 访谈：\textbf{美}国签证官解读学生签证获签要领 \\
\textbf{Translation:} Interview: \textbf{U.S.} visa officer interprets the essentials of student visa \\
 \\
$\mathcal{F}_T($美$) = $ 芥, 美, 界, 养, 镁, 每, 楣 (mustard, nice, world, support, magnesium, each, lintel) \\
$\mathcal{F}_K($美$) = $ 美 (US) \\
$\mathcal{F}_C($美$) = $ 美, 英, 香, 欧, 日, 澳, 俄, 荷, 德, 港, 华, 葡, 韩 (US, Britain, Hong Kong, Europe, Japan, Australia, Russian, Netherlands, Germany, Hong Kong, China, Portugal, Korean) \\

\midrule

\textbf{Input Chinese Text: } 陈嘉上《画皮》大换皮 凄\textbf{美}爱情赢得眼泪(图) \\
\textbf{Translation:} Chen Jia's "Painted Skin" changes skin, \textbf{poignant} love wins tears (photo) \\
\\
$\mathcal{F}_T($美$) = $ 芥, 美, 界, 养, 镁, 每, 楣 (mustard, nice, world, support, magnesium, each, lintel) \\
$\mathcal{F}_K($美$) = $ 美 (poignant) \\
$\mathcal{F}_C($美$) = $ 寞, 挚, 妙, 美, 腻, 酷, 烂, 凑, 坷, 凄, 惨, 悲, 慨 (lonely, sincere, wonderful, nice, greasy, cool, rotten, make up, bumpy, sad, awful, sad, sad) \\

\midrule

\textbf{Input Chinese Text: } 北京房协副秘书\textbf{长}陈志谈地产业诚信问题 \\
\textbf{Translation:} Chen Zhi, Deputy Secretary-\textbf{General} of the Beijing Housing Association, talks about the integrity of the real estate industry \\
\\
$\mathcal{F}_T($长$) = $ 氏, 氐, 掌, 涨, 长 (clan name, foundation, palm, rise, long) \\
$\mathcal{F}_K($长$) = $ 长 (general) \\
$\mathcal{F}_C($长$) = $ 长, 授, 卿, 员, 师, 委, 厅, 秘, 副, 顾, 官, 董 (general, professor, minister, member, teacher, committee, office, secretary, deputy, consultant, official, director) \\

\midrule

\textbf{Input Chinese Text: } IMF大幅下调明年全球经济增\textbf{长}预期 \\
\textbf{Translation:} IMF sharply lowered its forecast for global economic \textbf{growth} next year \\
\\
$\mathcal{F}_T($长$) = $ 氏, 氐, 掌, 涨, 长 (clan name, foundation, palm, rise, long) \\
$\mathcal{F}_K($长$) = $ 长, 加, 大 (growth, increase, increase) \\
$\mathcal{F}_C($长$) = $ 胀, 增, 膨, 速, 涨, 长, 加, 快 (swell, increase, inflate, speed, rise, grow, plus, fast) \\

\bottomrule
\end{tabular}
\end{table*}
\end{CJK*}

\newpage 
\subsection{English Adversarial Examples}
\label{appendix:examples}
\begin{CJK*}{UTF8}{gbsn}

\begin{table*}[htp!]\small \setlength{\tabcolsep}{7pt}
\centering
\caption{English Adversarial Examples Generated by \attack for BERT-based Classifier using $\mathcal{F}_T$.}
 \label{englishftexamples}
\begin{tabular}{p{13.8cm}}
\toprule Input (\textcolor{red}{red} = Modified character, \textbf{bold} = original character.) \\
\midrule

\textbf{Original English Text: } I went to AAA \textbf{for} their travel service. They could not help me at all with my trip to Belize. They have zilch information and resources. This is a prime destination of American tourists. I was disappointed. \\
\\
\textbf{Adversarial English Text: } I went to AAA \textcolor{red}{tor} their travel service. They could not help me at all with my trip to Belize. They have zilch information and resources. This is a prime destination of American tourists. I was disappointed. \\
\\
\textbf{Model Prediction: } 1-star (most negative) $\rightarrow$ 5-star (most positive) \\
\midrule

\textbf{Original English Text: } I called numerous times and noted that they are going to deliver at a work address between 9 am to 5 pm. They attempted delivery three times after 5 pm. I \textbf{got} ups to pick up my parcel and got it delivered on time. \\
\\
\textbf{Adversarial English Text: } I called numerous times and noted that they are going to deliver at a work address between 9 am to 5 pm. They attempted delivery three times after 5 pm. I \textcolor{red}{hot} ups to pick up my parcel and got it delivered on time. \\
\\
\textbf{Model Prediction: } 1-star (most negative) $\rightarrow$ 5-star (most positive) \\
\midrule

\textbf{Original English Text: } Mercedes does the \textbf{best} mani and pedi! You really have to go in at least once to see what I mean. \\
\\
\textbf{Adversarial English Text: } Mercedes does the \textcolor{red}{bet} mani and pedi! You really have to go in at least once to see what I mean. \\
\\
\textbf{Model Prediction: } 5-star (most positive) $\rightarrow$ 1-star (most negative) \\
\midrule

\textbf{Original English Text: } I was charged \$ 200 to \textbf{add} 6 lbs of Freon to my air conditioning. I went to amazon.com and 25 lbs cost \$ 120 including shipping. That should be approx \$ 29 \textbf{for} 6 lbs of Freon. So labor which was 20 min, transportation, and equipment up - keep for john, the service person who came, was \$ 171. I feel that's somewhat unreasonable. Just fair warning for the next customer. Update: after listening to my complaint, the owner offered to refund my payment. That was quite reasonable of them. Therefore, I switch my review to 4 stars. \\
\\
\textbf{Adversarial English Text: } I was charged \$ 200 to \textcolor{red}{ad} 6 lbs of Freon to my air conditioning. I went to amazon.com and 25 lbs cost \$ 120 including shipping. That should be approx \$ 29 \textcolor{red}{fog} 6 lbs of Freon. So labor which was 20 min, transportation, and equipment up - keep for john, the service person who came, was \$ 171. I feel that's somewhat unreasonable. Just fair warning for the next customer. Update: after listening to my complaint, the owner offered to refund my payment. That was quite reasonable of them. Therefore, I switch my review to 4 stars. \\
\\
\textbf{Model Prediction: } 4-star (positive) $\rightarrow$ 1-star (most negative) \\
\midrule

\textbf{Original English Text: } \textbf{Liked} how they were open late and also had happy hour specials after 10 pm. We really \textbf{liked} the bulgogi and korean prime kalbi. They were marinated very flavor-fully . the mushroom medley and sweet corn were also very good. Would definitely keep this place on my list of late night eats or when iia just craving korean barbecue. \\
\\
\textbf{Adversarial English Text: } \textcolor{red}{Lied} how they were open late and also had happy hour specials after 10 pm. We really \textcolor{red}{lied} the bulgogi and korean prime kalbi. They were marinated very flavor-fully . the mushroom medley and sweet corn were also very good. Would definitely keep this place on my list of late night eats or when iia just craving korean barbecue. \\
\\
\textbf{Model Prediction: } 4-star (positive) $\rightarrow$ 1-star (most negative) \\

\bottomrule
\end{tabular}
\end{table*}

\begin{table*}[htp!]\small \setlength{\tabcolsep}{7pt}
\centering
\caption{English Adversarial Examples Generated by \attack for BERT-based Classifier using $\mathcal{F}_K$.}
 \label{englishfkexamples}
\begin{tabular}{p{13.8cm}}
\toprule Input (\textcolor{red}{red} = Modified character, \textbf{bold} = original character.) \\
\midrule

\textbf{Original English Text: } Like the others below, I had a similar bad experience with this company. I also forgot to check here before I bought the living social deal. I am having some \textbf{issues} getting it refunded as well. Maid affordable was a no show, will not \textbf{call} back, and does not answer the phone or emails. Definitely take your business to someone else. \\
\\
\textbf{Adversarial English Text: } Like the others below, I had a similar bad experience with this company. I also forgot to check here before I bought the living social deal. I am having some \textcolor{red}{topic} getting it refunded as well. Maid affordable was a no show, will not \textcolor{red}{shout} back, and does not answer the phone or emails. Definitely take your business to someone else. \\
\\
\textbf{Model Prediction: } 1-star (most negative) $\rightarrow$ 5-star (most positive) \\
\midrule

\textbf{Original English Text: } Just another reason why I will never bank with chase.... so now you can't deposit any amount of cash without \textbf{showing} your id..... so much for just running to the bank quick. \\
\\
\textbf{Adversarial English Text: } Just another reason why I will never bank with chase.... so now you can't deposit any amount of cash without \textcolor{red}{usher} your id..... so much for just running to the bank quick. \\
\\
\textbf{Model Prediction: } 1-star (most negative) $\rightarrow$ 5-star (most positive) \\
\midrule

\textbf{Original English Text: } My 2017 camry got a check engine light and my car had a strong odor of gasoline after service closed, I asked the receptionist if there was anyway they could get me a rental and she said they were closed so she recommended me to come in bright and early at 7am on monday so they could look at my car so I told her I left for work at 6am cause I work in north scottsdale so I told her I didn't not want to drive my car to scottsdale and back because I was afraid my car would blow up or something from the strong odor of gasoline and she put me on hold to talk to a manager. When she came back on the phone she said her manager was going to get a hold of the rental manager to see if someone could come in tomorrow ( today now ) to get me a rental and I left my name and number and no one has \textbf{reached} out to me. It's great to know they don't care if their customer's car blows up on the freeway cause it's not a sale! Thanks avondale toyota you guys rock ! ! ! ! The dealership I work at teaches their receptionist to hand out rentals cause they know stuff like this happens, you guys might want to look into that ! \\
\\
\textbf{Adversarial English Text: } My 2017 camry got a check engine light and my car had a strong odor of gasoline after service closed, I asked the receptionist if there was anyway they could get me a rental and she said they were closed so she recommended me to come in bright and early at 7am on monday so they could look at my car so I told her I left for work at 6am cause I work in north scottsdale so I told her I didn't not want to drive my car to scottsdale and back because I was afraid my car would blow up or something from the strong odor of gasoline and she put me on hold to talk to a manager. When she came back on the phone she said her manager was going to get a hold of the rental manager to see if someone could come in tomorrow ( today now ) to get me a rental and I left my name and number and no one has \textcolor{red}{achieve} out to me. It's great to know they don't care if their customer's car blows up on the freeway cause it's not a sale! Thanks avondale toyota you guys rock ! ! ! ! The dealership I work at teaches their receptionist to hand out rentals cause they know stuff like this happens, you guys might want to look into that ! \\
\\
\textbf{Model Prediction: } 1-star (most negative) $\rightarrow$ 5-star (most positive) \\
\midrule

\textbf{Original English Text: } I called numerous times and noted that they are going to deliver at a work address between 9 am to 5 pm. They attempted delivery three \textbf{times} after 5 pm. I got up to pick up my parcel and got it delivered on time . \\
\\
\textbf{Adversarial English Text: } I called numerous times and noted that they are going to deliver at a work address between 9 am to 5 pm. They attempted delivery three \textcolor{red}{meter} after 5 pm. I got up to pick up my parcel and got it delivered on time . \\
\\
\textbf{Model Prediction: } 1-star (most negative) $\rightarrow$ 5-star (most positive) \\

\bottomrule
\end{tabular}
\end{table*}

\begin{table*}[htp!]\small \setlength{\tabcolsep}{7pt}
\centering
\caption{English Adversarial Examples Generated by \attack for BERT-based Classifier using $\mathcal{F}_C$.}
 \label{ansqasentexamples}
\begin{tabular}{p{13.8cm}}
\toprule Input (\textcolor{red}{red} = Modified character, \textbf{bold} = original character.) \\
\midrule

\textbf{Original English Text: } If you \textbf{think} Las Vegas is getting too white trash, don't go near here. This place is like a Steinbeck novel come to life. I kept expecting to see donkeys and chickens walking around. woo - pig - soooeeee this place is awful ! ! !\\
\\
\textbf{Adversarial English Text: } If you \textcolor{red}{senses} Las Vegas is getting too white trash, don't go near here. This place is like a Steinbeck novel come to life. I kept expecting to see donkeys and chickens walking around. woo - pig - soooeeee this place is awful ! ! ! \\
\\
\textbf{Model Prediction: } 1-star (most negative) $\rightarrow$ 5-star (most positive)\\
\midrule

\textbf{Original English Text: } My 2017 camry got a check engine light and my car had a strong odor of gasoline after service closed, I asked the receptionist if there was anyway they could get me a rental and she said they were closed so she recommended me to come in bright and early at 7am on monday so they could look at my car so I told her I left for work at 6am cause I work in north scottsdale so I told her I didn't not want to drive my car to scottsdale and back because I was \textbf{afraid} my car would blow up or something from the strong odor of gasoline and she put me on hold to talk to a manager. When she came back on the phone she said her manager was going to get a hold of the rental manager to see if someone could come in tomorrow ( today now ) to get me a rental and I left my name and number and no one has reached out to me. It's great to know they don't care if their customer's car blows up on the freeway cause it's not a sale ! Thanks avondale toyota you guys rock ! ! ! ! the dealership I work at teaches their receptionist to hand out rentals cause they know stuff like this happens, you guys might want to look into that ! \\
\\
\textbf{Adversarial English Text: } My 2017 camry got a check engine light and my car had a strong odor of gasoline after service closed, I asked the receptionist if there was anyway they could get me a rental and she said they were closed so she recommended me to come in bright and early at 7am on monday so they could look at my car so I told her I left for work at 6am cause I work in north scottsdale so I told her I didn't not want to drive my car to scottsdale and back because I was \textcolor{red}{worry} my car would blow up or something from the strong odor of gasoline and she put me on hold to talk to a manager. When she came back on the phone she said her manager was going to get a hold of the rental manager to see if someone could come in tomorrow ( today now ) to get me a rental and I left my name and number and no one has reached out to me. It's great to know they don't care if their customer's car blows up on the freeway cause it's not a sale ! Thanks avondale toyota you guys rock ! ! ! ! the dealership I work at teaches their receptionist to hand out rentals cause they know stuff like this happens, you guys might want to look into that ! \\
\\
\textbf{Model Prediction: } 1-star (most negative) $\rightarrow$ 5-star (most positive) \\
\midrule

\textbf{Original English Text: } I have used this company twice. The first time they were great. We spent over 5,000 for installation of a new ac unit on a rental property. Since they did an \textbf{excellent} job, we had them do a redesign of ac system in our home to improve the cooling in our house. It was one of the most frustrating customer service experiences I've had with a contractor in the 25 years I have lived in phoenix. They didn't complete the job in the time frame they promised. They damaged the faux ceiling in the kitchen, they drilled holes and didn't repair them in the bedroom. They left marks on the ceiling in the \textbf{living} room, where they marked to cut a hole and then didn't. Which told me they installers were not skilled or professional. After waiting for 2 months for them to repair the mistake in the kitchen, we gave up and paid to have it repaired. I heard a lot of promises, no solution. I would never use this contractor again. \\
\\
\textbf{Adversarial English Text: } I have used this company twice. The first time they were great. We spent over 5,000 for installation of a new ac unit on a rental property. Since they did an \textcolor{red}{exemplary} job, we had them do a redesign of ac system in our home to improve the cooling in our house. It was one of the most frustrating customer service experiences I've had with a contractor in the 25 years I have lived in phoenix. They didn't complete the job in the time frame they promised. They damaged the faux ceiling in the kitchen, they drilled holes and didn't repair them in the bedroom. They left marks on the ceiling in the \textcolor{red}{attic} room, where they marked to cut a hole and then didn't. Which told me they installers were not skilled or professional. After waiting for 2 months for them to repair the mistake in the kitchen, we gave up and paid to have it repaired. I heard a lot of promises, no solution. I would never use this contractor again. \\
\\
\textbf{Model Prediction: } 1-star (most negative) $\rightarrow$ 5-star (most positive) \\
\midrule

\textbf{Original English Text: } There's so many choices of food in Las vegas. Don't choose this place. It is no exaggeration that mcdonalds and arby's have better hash browns, eggs, and bacon. Missed items in the dishes we ordered. All around \textbf{disappointment} to the las vegas allure. \\
\\
\textbf{Adversarial English Text: } There's so many choices of food in Las vegas. Don't choose this place. It is no exaggeration that mcdonalds and arby's have better hash browns, eggs, and bacon. Missed items in the dishes we ordered. All around \textcolor{red}{sorrow} to the las vegas allure. \\
\\
\textbf{Model Prediction: } 1-star (most negative) $\rightarrow$ 5-star (most positive) \\

\bottomrule
\end{tabular}
\end{table*}

\begin{table*}[htp!]\small \setlength{\tabcolsep}{7pt}
\centering
\caption{English Adversarial Examples Generated by \attack  for BERT-based Classifier using all perturbation functions.}
 \label{ansqasentexamples}
\begin{tabular}{p{13.8cm}}
\toprule Input (\textcolor{red}{red} = Modified character, \textbf{bold} = original character.) \\
\midrule

\textbf{Original English Text: } I went to AAA for their travel service. They could not help me at all with my \textbf{trip} to Belize. They have zilch information and resources. This is a prime destination of American tourists. I was disappointed.\\
\\
\textbf{Adversarial English Text: } I went to AAA for their travel service. They could not help me at all with my \textcolor{red}{voyage} to Belize. They have zilch information and resources. This is a prime destination of American tourists. I was disappointed. \\
\\
\textbf{Model Prediction: } 1-star (most negative) $\rightarrow$ 5-star (most positive)\\
\midrule

\textbf{Original English Text: } My wife and I have been to this location multiple times, and have only had 1 \textbf{bad} experience where the people in the check out area were a little brain dead that day. (they told us that the rug we purchased wasn't in stock, then it was, then wasn't, then was again...) Other than that, we are always \textbf{helped} right away, and checking out goes quickly. They also have free self serve Starbucks coffee which I always help myself to. \\
\\
\textbf{Adversarial English Text: } My wife and I have been to this location multiple times, and have only had 1 \textcolor{red}{worst} experience where the people in the check out area were a little brain dead that day. (they told us that the rug we purchased wasn't in stock, then it was, then wasn't, then was again...) Other than that, we are always \textcolor{red}{served} right away, and checking out goes quickly. they also have free self serve Starbucks coffee which I always help myself to. \\
\\
\textbf{Model Prediction: } 4-star (positive) $\rightarrow$ 1-star (most negative) \\
\midrule

\textbf{Original English Text: } I love shopping at buffalo exchange but when it comes to selling I prefer selling to the phoenix location because the employees are a lot more genuine, there's less of a hipster pretentious vibe there, and I usually sell more there \textbf{too}. Not to mention the tempe location usually turns the music off at 8:30, which gives an unwanted feeling to their guests. I am giving two stars for the sake of finding things at all locations. Go phoenix location! \\
\\
\textbf{Adversarial English Text: } I love shopping at buffalo exchange but when it comes to selling I prefer selling to the phoenix location because the employees are a lot more genuine, there's less of a hipster pretentious vibe there, and I usually sell more there \textcolor{red}{anyway}. Not to mention the tempe location usually turns the music off at 8:30, which gives an unwanted feeling to their guests. I am giving two stars for the sake of finding things at all locations. Go phoenix location! \\
\\
\textbf{Model Prediction: } 2-star (negative) $\rightarrow$ 5-star (most positive) \\
\midrule

\textbf{Original English Text: } There' s so many choices of food in Las Vegas. Don't choose this place. It is no exaggeration that mcdonalds and arby's have \textbf{better} hash browns, eggs, and bacon. Missed items in the dishes we ordered. All around disappointment to the Las Vegas allure. \\
\\
\textbf{Adversarial English Text: } There's so many choices of food in Las Vegas. Don't choose this place. It is no exaggeration that mcdonalds and arby's have \textcolor{red}{delicious} hash browns, eggs, and bacon. Missed items in the dishes we ordered. All around disappointment to the Las Vegas allure. \\
\\
\textbf{Model Prediction: } 1-star (most negative) $\rightarrow$ 5-star (most positive) \\
\midrule

\textbf{Original English Text: } Not only is this place in my neighborhood, it is exactly what I'm looking for. I have pale skin, green eyes, and freckles yet I have been \textbf{cheated} out of having naturally red hair by mother nature!! Therefore I have been a fake redhead for at least a decade. You can imagine the cost and damage to my hair I have endured. Fringe has a new dye that is ammonia free! It's basically just a oil and water dying process! I've gone twice in a row and my hair has never been in such good condition. I'm paying the same amount for hair dying as my old salon except here I get a better cut and style and it's not frying my hair! Also Chanel (who dyes my hair) is a totally cool chic and always has interesting things to talk about! This is my new go to salon! \\
\\
\textbf{Adversarial English Text: } Not only is this place in my neighborhood, it is exactly what I'm looking for. I have pale skin, green eyes, and freckles yet I have been \textcolor{red}{humiliated} out of having naturally red hair by mother nature!! Therefore I have been a fake redhead for at least a decade. You can imagine the cost and damage to my hair I have endured. Fringe has a new dye that is ammonia free! It's basically just a oil and water dying process! I've gone twice in a row and my hair has never been in such good condition. I'm paying the same amount for hair dying as my old salon except here I get a better cut and style and it's not frying my hair! Also Chanel (who dyes my hair) is a totally cool chic and always has interesting things to talk about! This is my new go to salon! \\
\\
\textbf{Model Prediction: } 5-star (most positive) $\rightarrow$ 1-star (most negative) \\

\bottomrule
\end{tabular}
\end{table*}

\begin{table*}[htp!]\small \setlength{\tabcolsep}{7pt}
\centering
\caption{English Adversarial Examples Generated by \attack  for BERT-based Classifier on SNLI Dataset using all perturbation functions.}
 \label{ansqasentexamples}
\begin{tabular}{p{13.8cm}}
\toprule Input (\textcolor{red}{red} = Modified character, \textbf{bold} = original character.) \\
\midrule

\textbf{Original Premise: } Four boys are about to be \textbf{hit} by an approaching wave.\\
\textbf{Adversarial Premise: } Four boys are about to be \textcolor{red}{smashed} by an approaching wave.\\
\textbf{Hypothesis: } The wave missed the boys. \\
\\
\textbf{Model Prediction: } contradiction $\rightarrow$ entailment\\
\midrule

\textbf{Original Premise: } A yellow race car \textbf{sliding} through a corner as spectators watch.\\
\textbf{Adversarial Premise: } A yellow race car \textcolor{red}{slipping} through a corner as spectators watch.\\
\textbf{Hypothesis: } A NASCAR is being watched. \\
\\
\textbf{Model Prediction: } neutral $\rightarrow$ entailment\\
\midrule

\textbf{Original Premise: } A group of people on the bark, brightly lighten street, while one man with a gray hat holds a large colorful sign with \textbf{arrows}.\\
\textbf{Adversarial Premise: } A group of people on the bark, brightly lighten street, while one man with a gray hat holds a large colorful sign with \textcolor{red}{swords}.\\
\textbf{Hypothesis: } The people are walking down the street. \\
\\
\textbf{Model Prediction: } entailment $\rightarrow$ contradiction\\
\midrule

\textbf{Original Premise: } A man takes a \textbf{drink} in the doorway of a home.\\
\textbf{Adversarial Premise: } A man takes a \textcolor{red}{drinking} in the doorway of a home.\\
\textbf{Hypothesis: } A man is looking out onto his front lawn from the doorway of his home. \\
\\
\textbf{Model Prediction: } neutral $\rightarrow$ contradiction\\
\midrule

\textbf{Original Premise: } A dog \textbf{attacking} a man wearing protective gear.\\
\textbf{Adversarial Premise: } A dog \textcolor{red}{hurting} a man wearing protective gear.\\
\textbf{Hypothesis: } He was training a police dog. \\
\\
\textbf{Model Prediction: } neutral $\rightarrow$ entailment\\
\midrule

\textbf{Original Premise: } A white man in a \textbf{red} shirt riding a bike.\\
\textbf{Adversarial Premise: } A white man in a \textcolor{red}{golden} shirt riding a bike.\\
\textbf{Hypothesis: } An old guy wears a shirt on a bike. \\
\\
\textbf{Model Prediction: } neutral $\rightarrow$ entailment\\
\midrule

\textbf{Original Premise: } A child in a blue and white striped shirt crosses his arms and smiles while standing on red carpeted \textbf{stairs}.\\
\textbf{Adversarial Premise: } A child in a blue and white striped shirt crosses his arms and smiles while standing on red carpeted \textcolor{red}{terraces}.\\
\textbf{Hypothesis: } A child is smiling as he watches a clown. \\
\\
\textbf{Model Prediction: } neutral $\rightarrow$ contradiction\\
\midrule

\textbf{Original Premise: } This man, with a red \& white shirt has \textbf{water} bottles on this white truck.\\
\textbf{Adversarial Premise: } This man, with a red \& white shirt has \textcolor{red}{beer} bottles on this white truck.\\
\textbf{Hypothesis: } The guy has bottles on the truck for me. \\
\\
\textbf{Model Prediction: } neutral $\rightarrow$ entailment\\
\midrule

\textbf{Original Premise: } Three people are riding a carriage \textbf{pulled} by four horses.\\
\textbf{Adversarial Premise: } Three people are riding a carriage \textcolor{red}{hauled} by four horses.\\
\textbf{Hypothesis: } The oxen are pulling the carriage. \\
\\
\textbf{Model Prediction: } contradiction $\rightarrow$ entailment\\

\bottomrule
\end{tabular}
\end{table*}

\newpage 
\subsection{Chinese Adversarial Examples}

\begin{table*}[htp!]\small \setlength{\tabcolsep}{7pt}
\centering
\caption{Chinese Adversarial Examples Generated by \attack for BERT-based Classifier on THUNews Dataset using $\mathcal{F}_T$.}
 \label{ansqasentexamples}
\begin{tabular}{p{13.8cm}}
\toprule Input (\textcolor{red}{red} = Modified character, \textbf{bold}=original character.) \\
\midrule

\textbf{Original Chinese Text: } 高露\textbf{洁}新品专效抗敏牙膏解决牙齿过敏 \\
\textbf{Translation:} Gaolu\textbf{jie}'s new anti-hypersensitive toothpaste solves tooth hypersensitivity \\
 \\
\textbf{Adversarial Chinese Text: } 高露\textcolor{red}{吉}新品专效抗敏牙膏解决牙齿过敏 \\
\textbf{Translation: } Gaolu\textcolor{red}{ji}'s new anti-hypersensitive toothpaste solves tooth hypersensitivity \\
\\
\textbf{Model Prediction: } Fashion News （时尚新闻）$\rightarrow$  Entertainment News （娱乐新闻） \\
\midrule

\textbf{Original Chinese Text: } 组图：09巴黎高级定制秀最有看点8场\textbf{次} \\
\textbf{Translation:} Photos: 8 \textbf{highlights} of 09 Paris Haute Couture Show  \\
 \\
\textbf{Adversarial Chinese Text: } 组图：09巴黎高级定制秀最有看点8场\textcolor{red}{炊} \\
\textbf{Translation: } Photos: 8 \textcolor{red}{cooking sessions} of 09 Paris Haute Couture Show \\
 \\
\textbf{Model Prediction: } Fashion News （时尚新闻）$\rightarrow$  Entertainment News （娱乐新闻） \\
\midrule

\textbf{Original Chinese Text: } \textbf{今}秋男友新标准打造新时代型男  \\
\textbf{Translation:} New standards for boyfriends in \textbf{this} autumn to create a new era of men  \\
 \\
\textbf{Adversarial Chinese Text: } \textcolor{red}{金}秋男友新标准打造新时代型男
 \\
\textbf{Translation: } New standards for boyfriends in \textcolor{red}{golden} autumn to create a new era of men \\
 \\
\textbf{Model Prediction: } Fashion News （时尚新闻）$\rightarrow$  Entertainment News （娱乐新闻）\\
\midrule

\textbf{Original Chinese Text: } \textbf{据}称台联党可能下令赖幸媛辞去陆委会主委  \\
\textbf{Translation:} \textbf{It is} said that the Taiwan Union Party may order Lai Xingyuan to resign as chairman of the MAC \\
 \\
\textbf{Adversarial Chinese Text: } \textcolor{red}{剧}称台联党可能下令赖幸媛辞去陆委会主委 \\
\textbf{Translation: } \textcolor{red}{The drama} said that the Taiwan Union Party may order Lai Xingyuan to resign as chairman of the MAC \\
 \\
\textbf{Model Prediction: } Politics news （时政新闻） $\rightarrow$ Entertainment News （娱乐新闻） \\
\midrule

\textbf{Original Chinese Text: } \textbf{猛}犸象$80\%$基因组破译完成史前巨兽有望复活  \\
\textbf{Translation:} \textbf{Mammoth} $80\%$ genome deciphered complete prehistoric behemoth is expected to be resurrected  \\
 \\
\textbf{Adversarial Chinese Text: } \textcolor{red}{孟}犸象$80\%$基因组破译完成史前巨兽有望复活 \\
\textbf{Translation: } \textcolor{red}{Mammoth} $80\%$ genome deciphered complete prehistoric behemoth is expected to be resurrected  \\
 \\
\textbf{Model Prediction: } Technology News （科技新闻） $\rightarrow$  Entertainment News （娱乐新闻）\\

\bottomrule
\end{tabular}
\end{table*}

\begin{table*}[htp!]\small \setlength{\tabcolsep}{7pt}
\centering
\caption{Chinese Adversarial Examples Generated by \attack for BERT-based Classifier on THUNews Dataset using $\mathcal{F}_K$.}
 \label{ansqasentexamples}
\begin{tabular}{p{13.8cm}}
\toprule Input (\textcolor{red}{red} = Modified character, \textbf{bold}=original character.) \\
\midrule

\textbf{Original Chinese Text: } 手袋进阶论：职场之路的\textbf{秘}密奠基石（组图） \\
\textbf{Translation:} Handbag progression theory: the \textbf{secret} cornerstone of the road to the workplace (photo) \\
 \\
\textbf{Adversarial Chinese Text: } 手袋进阶论：职场之路的\textcolor{red}{机}密奠基石（组图） \\
\textbf{Translation: } Handbag progression theory: the \textcolor{red}{confidential} cornerstone of the road to the workplace (photo) \\
\\
\textbf{Model Prediction: } Fashion News （时尚新闻）$\rightarrow$  Technology News （科技新闻） \\
\midrule

\textbf{Original Chinese Text: } 中国银联发布十一黄金周用卡提\textbf{示} \\
\textbf{Translation:} China UnionPay releases card \textbf{tips} for Golden Week.  \\
 \\
\textbf{Adversarial Chinese Text: } 中国银联发布十一黄金周用卡提\textcolor{red}{醒} \\
\textbf{Translation: } China UnionPay releases card \textcolor{red}{reminders} for Golden Week. \\
 \\
\textbf{Model Prediction: } Financial and economic news （财经新闻）$\rightarrow$  Technology News （科技新闻） \\
\midrule

\textbf{Original Chinese Text: } 买卖红木都是一项\textbf{风}险活 \\
\textbf{Translation:} Buying and selling mahogany is a \textbf{risky} business. \\
 \\
\textbf{Adversarial Chinese Text: } 买卖红木都是一项\textcolor{red}{危}险活
 \\
\textbf{Translation: } Buying and selling mahogany is a \textcolor{red}{dangerous} business. \\
 \\
\textbf{Model Prediction: } Financial and economic news （财经新闻）$\rightarrow$  Home News （家居新闻）\\
\midrule

\textbf{Original Chinese Text: } 信用卡利润猛涨风险容忍度提\textbf{高} \\
\textbf{Translation:} Credit card profits soar with \textbf{increased} risk tolerance. \\
 \\
\textbf{Adversarial Chinese Text: } 信用卡利润猛涨风险容忍度提\textcolor{red}{升} \\
\textbf{Translation: } Credit card profits soar with \textcolor{red}{increased} risk tolerance. \\
 \\
\textbf{Model Prediction: } Financial and economic news （财经新闻） $\rightarrow$ Stock News （股票新闻） \\
\midrule

\textbf{Original Chinese Text: } 黎振伟：不同的\textbf{城}市有着各自的发展模式 \\
\textbf{Translation:} Zhenwei Li: Different \textbf{cities} have their own development models. \\
 \\
\textbf{Adversarial Chinese Text: } 黎振伟：不同的\textcolor{red}{都}市有着各自的发展模式 \\
\textbf{Translation: } Zhenwei Li: Different \textcolor{red}{cities} have their own development models. \\
 \\
\textbf{Model Prediction: } Real Estate News （房产新闻） $\rightarrow$  Technology News （科技新闻）\\
\midrule

\textbf{Original Chinese Text: } 韩国航空\textbf{试}验中心揭秘：战斗机被冰冻住测试 \\
\textbf{Translation:} South Korea's aviation \textbf{experiment} center revealed: fighter jets were frozen in the test. \\
 \\
\textbf{Adversarial Chinese Text: } 韩国航空\textcolor{red}{检}验中心揭秘：战斗机被冰冻住测试 \\
\textbf{Translation: } South Korea's aviation \textcolor{red}{test} center revealed: fighter jets were frozen in the test. \\
 \\
\textbf{Model Prediction: } Technology News （科技新闻） $\rightarrow$  Current Affairs News （时政新闻）\\

\bottomrule
\end{tabular}
\end{table*}

\begin{table*}[htp!]\small \setlength{\tabcolsep}{7pt}
\centering
\caption{Chinese Adversarial Examples Generated by \attack for BERT-based Classifier on THUNews Dataset using $\mathcal{F}_C$.}
 \label{ansqasentexamples}
\begin{tabular}{p{13.8cm}}
\toprule Input (\textcolor{red}{red} = Modified character, \textbf{bold}=original character.) \\
\midrule

\textbf{Original Chinese Text: } 高露\textbf{洁}新品专效抗敏牙膏解决牙齿过敏 \\
\textbf{Translation:} Gaolu\textbf{jie}'s new anti-hypersensitive toothpaste solves tooth hypersensitivity \\
 \\
\textbf{Adversarial Chinese Text: } 高露\textcolor{red}{婕}新品专效抗敏牙膏解决牙齿过敏 \\
\textbf{Translation: } Gaolu\textcolor{red}{jie}'s new anti-hypersensitive toothpaste solves tooth hypersensitivity \\
\\
\textbf{Model Prediction: } Fashion News （时尚新闻）$\rightarrow$  Entertainment News （娱乐新闻） \\
\midrule

\textbf{Original Chinese Text: } 实\textbf{录}：张瑜阿穆隆王睿做客聊新片《八十一格》 \\
\textbf{Translation:} \textbf{Record}: Zhang Yu, Amulon and Wang Rui as a guest to talk about the new film "Eighty-one Patterns" \\
 \\
\textbf{Adversarial Chinese Text: } 实\textcolor{red}{摄}：张瑜阿穆隆王睿做客聊新片《八十一格》 \\
\textbf{Translation: } \textcolor{red}{Record}: Zhang Yu, Amulon and Wang Rui as a guest to talk about the new film "Eighty-one Patterns" \\
\\
\textbf{Model Prediction: } Entertainment News （娱乐新闻） $\rightarrow$ Technology News （科技新闻） \\
\midrule

\textbf{Original Chinese Text: } 聚\textbf{焦}信用卡全额罚息：欠款 $44.6$ 元生千元利息 \\
\textbf{Translation:} \textbf{Focus} on credit card full penalty interest: RMB $44.6$ arrears generate interest of RMB $1,000$ \\
 \\
\textbf{Adversarial Chinese Text: } 聚\textcolor{red}{盯}信用卡全额罚息：欠款 $44.6$ 元生千元利息 \\
\textbf{Translation: } \textcolor{red}{Focus} on credit card full penalty interest: RMB $44.6$ arrears generate interest of RMB $1,000$ \\
 \\
\textbf{Model Prediction: } Financial and economic news （财经新闻） $\rightarrow$ Technology News （科技新闻） \\
\midrule

\textbf{Original Chinese Text: } 研究发现4000万年前\textbf{鲸}鱼长有4条腿（图） \\
\textbf{Translation:} Research found that \textbf{whales} had 4 legs 40 million years ago (photo) \\
 \\
\textbf{Adversarial Chinese Text: } 研究发现4000万年前\textcolor{red}{鲤}鱼长有4条腿（图） \\
\textbf{Translation: } Research found that \textcolor{red}{carp} had 4 legs 40 million years ago (photo) \\
 \\
\textbf{Model Prediction: } Technology News （科技新闻） $\rightarrow$ Social News （社会新闻） \\
\midrule

\textbf{Original Chinese Text: } 澳门博彩业后何鸿时代\textbf{猜}想 \\
\textbf{Translation:} Post-Ho Hong Era \textbf{Conjecture} in Macau's Gaming Industry \\
 \\
\textbf{Adversarial Chinese Text: } 澳门博彩业后何鸿时代\textcolor{red}{预}想 \\
\textbf{Translation: } Post-Ho Hong Era \textcolor{red}{Prediction} in Macau's Gaming Industry \\
 \\
\textbf{Model Prediction: }  Stock news （股票新闻） $\rightarrow$  Technology News （科技新闻）\\

\bottomrule
\end{tabular}
\end{table*}

\begin{table*}[htp!]\small \setlength{\tabcolsep}{7pt}
\centering
\caption{Chinese Adversarial Examples Generated by \attack for BERT-based Classifier on THUNews Dataset using all perturbation functions.}
 \label{ansqasentexamples}
\begin{tabular}{p{13.8cm}}
\toprule Input (\textcolor{red}{red} = Modified character, \textbf{bold}=original character.) \\
\midrule

\textbf{Original Chinese Text: } 对 话 王 辉 \textbf{灏}： 海 归 创 业 面 临 的 困 难 （图） \\
\textbf{Translation:} Dialogue with Wang \textbf{Huihao}: Difficulties faced by overseas returnees in starting a business (photo)  \\
\\
\textbf{Adversarial Chinese Text: } 对 话 王 辉 \textcolor{red}{耀}： 海 归 创 业 面 临 的 困 难 （图） \\
\textbf{Translation: } Dialogue with Wang \textcolor{red}{Huiyao}: Difficulties faced by overseas returnees in starting a business (photo) \\
\\
\textbf{Model Prediction: } Education News （教育新闻） $\rightarrow$ Entertainment News （娱乐新闻）\\
\midrule

\textbf{Original Chinese Text: } 拿 \textbf{什}么 能 吸 引 你 ： 我 们 的 海 外 学 子 ？ \\
\textbf{Translation:} \textbf{What} can attract you: our overseas students? \\
 \\
\textbf{Adversarial Chinese Text: } 拿 \textcolor{red}{甚}么 能 吸 引 你 ： 我 们 的 海 外 学 子 ？
 \\
\textbf{Translation: } \textcolor{red}{What} can attract you: our overseas students? \\
\\
\textbf{Model Prediction: } Education News （教育新闻） $\rightarrow$ Entertainment News （娱乐新闻） \\
\midrule

\textbf{Original Chinese Text: } 独 家 对 话 \textbf{冯}小 刚： 多 个 观 众 挺 难 少 点 观 众  挺 容 易 \\
\textbf{Translation:} Exclusive dialogue with \textbf{Feng} Xiaogang: It's difficult for multiple audiences, and it's easy for less audiences  \\
 \\
\textbf{Adversarial Chinese Text: } 独 家 对 话 \textcolor{red}{郜}小 刚 ： 多 个 观 众 挺 难 少 点 观 众 挺 容 易
 \\
\textbf{Translation: } Exclusive dialogue with \textcolor{red}{Gao} Xiaogang: It's difficult for multiple audiences, and it's easy for less audiences \\
 \\
\textbf{Model Prediction: } Entertainment News （娱乐新闻） $\rightarrow$ Sports News （体育新闻） \\
\midrule

\textbf{Original Chinese Text: } 高 露 洁 新 品 专 效 抗 敏 牙 膏 解 决 牙 \textbf{齿}过 敏  \\
\textbf{Translation:} Gaolujie's new anti-hypersensitive toothpaste solves \textbf{tooth} hypersensitivity  \\
 \\
\textbf{Adversarial Chinese Text: } 高 露 洁 新 品 专 效 抗 敏 牙 膏 解 决 牙 \textcolor{red}{苔}过 敏
 \\
\textbf{Translation: } Gaolujie's new anti-hypersensitive toothpaste solves \textcolor{red}{tooth coating} hypersensitivity \\
 \\
\textbf{Model Prediction: } Fashion News （时尚新闻）$\rightarrow$  Entertainment News （娱乐新闻）\\
\midrule

\textbf{Original Chinese Text: } 2010 艺 术 品 秋 拍 上 演 六 \textbf{宗}最  \\
\textbf{Translation:} Six \textbf{most} of the 2010 art autumn auctions \\
 \\
\textbf{Adversarial Chinese Text: } 2010 艺 术 品 秋 拍 上 演 六 \textcolor{red}{综}最
 \\
\textbf{Translation: } Six \textcolor{red}{most comprehensive} of the 2010 art autumn auctions \\
 \\
\textbf{Model Prediction: } Financial and economic news （财经新闻） $\rightarrow$ Entertainment News （娱乐新闻） \\
\midrule

\textbf{Original Chinese Text: } 英属小岛发现罕见蓝色龙\textbf{虾}（组图）  \\
\textbf{Translation:} Rare blue \textbf{lobster} found on British island (photo)  \\
 \\
\textbf{Adversarial Chinese Text: } 英属小岛发现罕见蓝色龙\textcolor{red}{鳖}（组图） \\
\textbf{Translation: } Rare blue \textcolor{red}{turtle} found on British island (photo)  \\
 \\
\textbf{Model Prediction: } Technology News （科技新闻） $\rightarrow$  Social News （社会新闻）\\

\bottomrule
\end{tabular}
\end{table*}

\begin{table*}[htp!]\small \setlength{\tabcolsep}{7pt}
\centering
\caption{Chinese Adversarial Examples Generated by \attack for BERT-based Classifier on Wechat Finance Dataset using all perturbation functions.}
 \label{wechatexample}
\begin{tabular}{p{13.8cm}}
\toprule Input (\textcolor{red}{red} = Modified character, \textbf{bold}=original character.) \\
\midrule

\textbf{Original Chinese Text: } 翻倍网分享财富资产管理资讯知识技巧。关注信托、融资租赁、期\textbf{货}保险、私人银行等领域最新信息。 \\
\textbf{Translation: } Fanbei.com shares wealth and asset management information knowledge and skills. Pay attention to the latest information in the fields of trust, financial leasing, \textbf{futures} insurance, and private banking. \\
 \\
\textbf{Adversarial Chinese Text: } 翻倍网分享财富资产管理资讯知识技巧。关注信托、融资租赁、期\textcolor{red}{祸}保险、私人银行等领域最新信息。
 \\
\textbf{Translation: } Fanbei.com shares wealth and asset management information knowledge and skills. Pay attention to the latest information in the fields of trust, financial leasing, \textcolor{red}{accident} insurance, and private banking. \\
\\
\textbf{Model Prediction: } Comprehensive （综合） $\rightarrow$ Bank（银行）\\
\midrule

\textbf{Original Chinese Text: } 温泉邮政\textbf{支}局提供邮政服务、个性化邮票订制、快递小包上门取件、邮件查询。 \\
\textbf{Translation:} The {Post Office} at Hot Spring \textbf{Branch} provides postal services, personalized stamp ordering, home delivery of small parcels, and mail inquiries. \\
 \\
\textbf{Adversarial Chinese Text: } 温泉邮政\textcolor{red}{驿}局提供邮政服务、个性化邮票订制、快递小包上门取件、邮件查询。
 \\
\textbf{Translation: } The Hot Spring \textcolor{red}{Post Office} provides postal services, personalized stamp ordering, home delivery of small parcels, and mail inquiries. \\
 \\
\textbf{Model Prediction: } Bank（银行） $\rightarrow$ Insurance（保险）\\
\midrule

\textbf{Original Chinese Text: } \textbf{中}融华创（北京）基金有限公司（简称：中融华创）成立于 2012 年 3 月 29 日。总部设立在首都北京，公司在国家发展改革委员会登记备案，由中国证券投资基金协会颁发金融牌照。
\\
\textbf{Translation:} \textbf{Zhongrong} Huachuang (Beijing) Fund Co., Ltd. (abbreviated as Zhongrong Huachuang) was established on March 29, 2012. Headquartered in the capital, Beijing, the company is registered with the National Development and Reform Commission, and is a legal financial institution that is issued a financial license by the Securities Investment Fund Association of China.
\\
\\
\textbf{Adversarial Chinese Text: } \textcolor{red}{申}融华创（北京）基金有限公司（简称：中融华创）成立于 2012 年 3 月 29 日。总部设立在首都北京，公司在国家发展改革委员会登记备案，由中国证券投资基金协会颁发金融牌照。
\\
\textbf{Translation: } \textcolor{red}{Shenrong} Huachuang (Beijing) Fund Co., Ltd. (abbreviated as Zhongrong Huachuang) was established on March 29, 2012. Headquartered in the capital, Beijing, the company is registered with the National Development and Reform Commission, and is a legal financial institution that is issued a financial license by the Securities Investment Fund Association of China.
\\
\\
\textbf{Model Prediction: } Fund （基金） $\rightarrow$ Comprehensive （综合）\\
\midrule

\textbf{Original Chinese Text: } 期\textbf{货}行业风起云涌，期市行情熟悉万变。交易帮玩转交易，携手众多期货高手，让交易更简单！ \\
\textbf{Translation:} The \textbf{futures} industry is surging, and the futures market is familiar with ever-changing conditions. Trading helps fun trading, and join hands with many futures experts to make trading easier!  \\
 \\
\textbf{Adversarial Chinese Text: } 期\textcolor{red}{券}行业风起云涌，期市行情熟悉万变。交易帮玩转交易，携手众多期货高手，让交易更简单！
 \\
\textbf{Translation: } The \textcolor{red}{futures bond} industry is surging, and the futures market is familiar with ever-changing conditions. Trading helps fun trading, and join hands with many futures experts to make trading easier! \\
 \\
\textbf{Model Prediction: } Futures （期货）$\rightarrow$  Comprehensive （综合）\\
\midrule

\textbf{Original Chinese Text: } 瑞\textbf{倪}资本专注于股权投资、证券投资及衍生品研究等领域，业务涵盖一、二级市场，包括天使投资以及对冲型、权益类与固定收益类证券投资。   \\
\textbf{Translation:} Rui\textbf{ni} Capital focuses on equity investment, securities investment and derivatives research and other fields. Its business covers primary and secondary markets, including angel investment and hedging, equity and fixed income securities investment. \\
 \\
\textbf{Adversarial Chinese Text: } 瑞\textcolor{red}{券}资本专注于股权投资、证投资及衍生品研究等领域，业务涵盖一、二级市场，包括天使投资以及对冲型、权益类与固定收益类证券投资。
 \\
\textbf{Translation: } Rui\textcolor{red}{quan} Capital focuses on equity investment, securities investment and derivatives research and other fields. Its business covers primary and secondary markets, including angel investment and hedging, equity and fixed income securities investment. \\
 \\
\textbf{Model Prediction: } Comprehensive （综合）$\rightarrow$  Securities （证券）\\

\bottomrule
\end{tabular}
\end{table*}

\end{CJK*}

\end{document}